\documentclass{article}

\usepackage[main, final]{neurips_2026}

\usepackage[utf8]{inputenc}
\usepackage[T1]{fontenc}
\usepackage{hyperref}
\usepackage{url}
\usepackage{booktabs}
\usepackage{amsfonts}
\usepackage{amsmath}
\usepackage{amssymb}
\usepackage{amsthm}
\usepackage{nicefrac}
\usepackage{microtype}
\usepackage{xcolor}
\usepackage{graphicx}
\usepackage{subcaption}
\usepackage[export]{adjustbox}
\newtheorem{proposition}{Proposition}

\title{Linear Generator Networks for Nonlinear Partial Differential Equations}

\author{%
  Shafayeth Jamil \\
  Ming Hsieh Department of Electrical and Computer Engineering \\
  University of Southern California \\
  Los Angeles, CA 90089 \\
  \texttt{sjamil@usc.edu} \\
  \And
  Rehan Kapadia \\
  Ming Hsieh Department of Electrical and Computer Engineering \\
  University of Southern California \\
  Los Angeles, CA 90089 \\
  \texttt{rkapadia@usc.edu} \\
}

\begin{document}
\raggedbottom
\setlength{\textfloatsep}{10pt plus 2pt minus 2pt}
\maketitle

\begin{abstract}
Most physical systems are governed by nonlinear partial differential equations, yet the analytical tools that make linear systems transparent, such as eigenspectra, dispersion relations and modal decomposition, remain unavailable in nonlinear settings. We introduce Linear Generator Network--Koopman (LGN-KM), a neural operator that learns a nonlinear lift and a continuous-time linear generator for the lifted dynamics through the decomposition $L_k = S - D_k$, where $S$ is skew-symmetric (conservative coupling) and $D_k$ is positive-definite diagonal (mode-selective dissipation). The decomposition guarantees stability by construction and exposes the eigenstructure directly, making linear-systems analysis available on a nonlinear PDE. On 2D Navier--Stokes turbulence, the generator yields a complete multi-branch dispersion relation and models trained independently at viscosities $100\times$ apart learn the same coupling spectrum ($R^2 = 0.87$). The same architecture shows the corresponding diffusive damping on FitzHugh--Nagumo reaction-diffusion, and for pulse propagation in optical fibers, the oscillation frequencies of the learned generator match the fiber's dispersion relation, giving the group-velocity dispersion to within $3.1\%$ with no fitted constant. The architecture additionally enables long-horizon stability, $O(1)$ continuous-time evaluation, and physics-informed cross-regime model transfer.
\end{abstract}

\begin{figure}[!t]
\centering
\includegraphics[width=\linewidth]{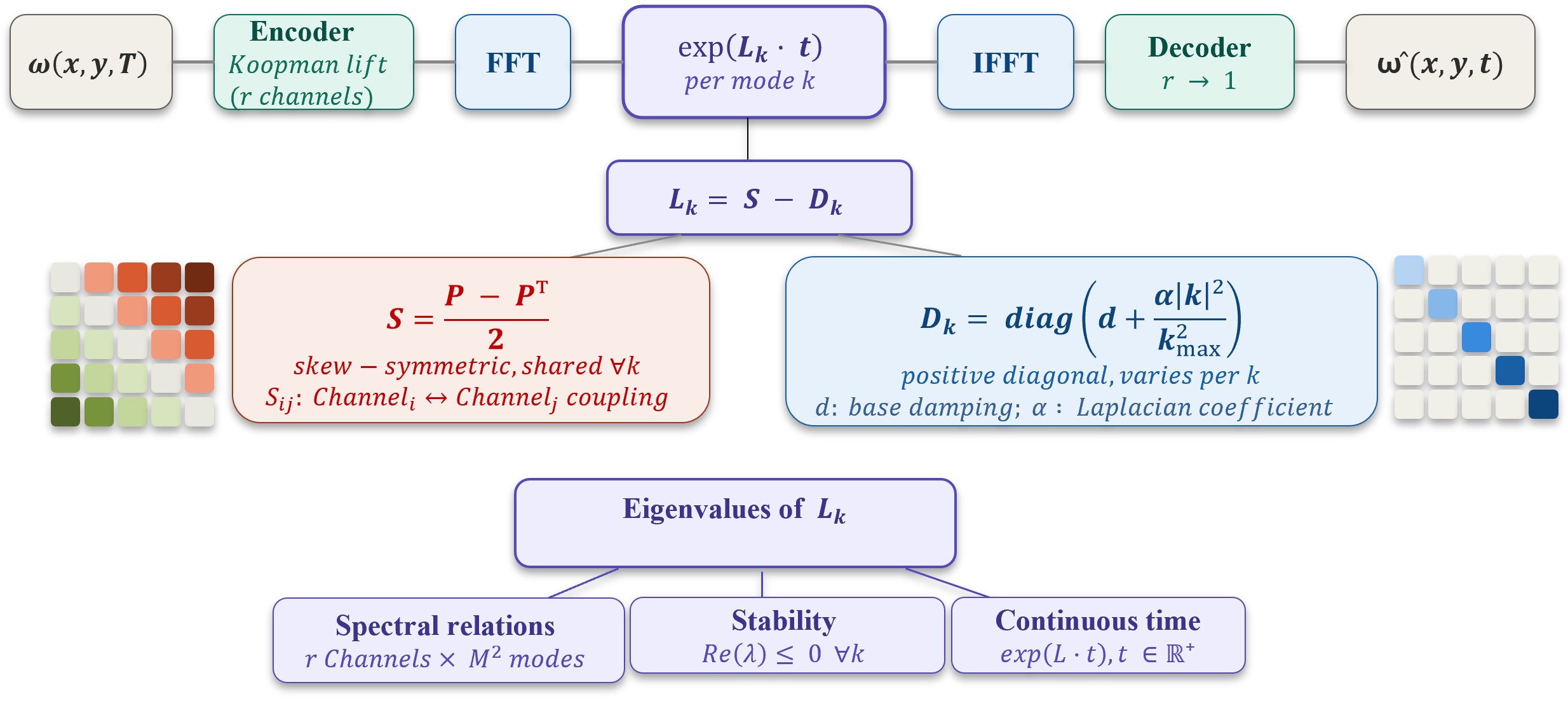}
\caption{LGN-KM architecture. Input nonlinear data is lifted to $r$ channels via a spectral encoder, transformed to Fourier space, and propagated by the matrix exponential $\exp(L_k\,t)$ of a per-mode generator $L_k = S - D_k$. The skew-symmetric component $S = (P - P^{\top})/2$ is shared across all wavevectors. The positive-definite diagonal $D_k = \mathrm{diag}\!\left(d + \alpha |k|^{2}/k_{\max}^{2}\right)$ varies per wavevector and encodes mode-selective dissipation with learned base damping $d$ and Laplacian coefficient $\alpha$. The eigenvalues of $L_k$ are guaranteed to have non-positive real parts, enabling a complete spectral characterization of the dynamics across $r \times M^{2}$ eigenvalues, long-horizon stability, and continuous-time evaluation at arbitrary $t$ via a single matrix exponentiation.}
\label{fig:architecture}
\end{figure}

\section{Introduction}

Most physical systems are nonlinear, yet many of our most powerful tools for understanding dynamics assume linearity. Eigenvalue analysis, modal decomposition, stability criteria, dispersion relations, and transfer functions all rely on a linear generator; once superposition fails, these tools no longer apply globally. For nonlinear systems, such information is typically available only through local linearization around specific operating points or through problem-specific theory. The result is a persistent asymmetry where linear systems are often analyzed directly, while nonlinear systems are more often explored through simulation.

Neural operators [1, 2, 3] have narrowed the computational side of this gap. They learn solution operators for PDEs from data, often delivering high accuracy and large speedups over classical solvers. But they do not usually recover an interpretable generator. Their outputs map inputs to future states, without exposing the modal structure, coupling mechanisms, or stability properties that make linear systems analytically transparent. Prediction has improved but diagnosis has not.

The Koopman formalism offers a path across this gap. For systems admitting a finite-dimensional invariant subspace, the nonlinear dynamics lift to a linear evolution $z(t) = \exp(L \cdot t)\, z_0$, where the generator $L$ carries the full analytical structure of a linear system applied to the original nonlinear dynamics. Existing deep Koopman methods [4, 5] learn the discrete-time operator $K = \exp(L)$ rather than $L$ itself, forfeiting direct spectral access. We introduce LGN-KM (Linear Generator Network---Koopman), which learns $L$ through a structured decomposition $L_k = S - D_k$ where $S$ is skew-symmetric and shared across Fourier modes (conservative inter-modal coupling), and $D_k$ is positive-definite diagonal with learned wavenumber dependence (dissipation). The $S{-}D$ decomposition was introduced in LGN for linear system identification [6]; here we extend it to nonlinear PDEs through Koopman lifting with per-Fourier-mode generators. Recent works learn continuous-time Koopman generators [7, 20] or constrain diagonal generators for stable rollouts [8], but none combines a stability guarantee with inter-mode coupling or reads interpretable physics from the generator.

The learned generator makes linear-systems analysis available for nonlinear PDEs. We demonstrate on 2D Navier--Stokes turbulence, where the learned generator yields a complete multi-branch dispersion relation with wavenumber-dependent eigenvalues, and a learnable dissipation exponent rises to 2.2--2.4 at the paper's horizons. The same architecture, with no modification, shows the analogous diffusive damping on 2D FitzHugh--Nagumo reaction-diffusion (Appendix~\ref{app:fhn}). This suggests that the $S{-}D$ decomposition captures structure shared across PDE families rather than a Navier--Stokes-specific fit. Models trained independently at different viscosities show gauge invariance despite learning in different latent coordinates. The architectural constraint simultaneously guarantees long-horizon stability where unconstrained operators diverge and enables $O(1)$ continuous-time evaluation. The universal coupling structure transfers across flow regimes with improved data efficiency. These capabilities come at a deliberate trade-off in one-step prediction accuracy, inherent to restricting the model class to stable, interpretable generators.\footnote{Code and checkpoints: \url{https://github.com/ShafayethJamil/Linear-Generator-Network-Koopman}}

\section{Method}

\subsection{Linear generators in the latent space}
Consider a dynamical system $dx/dt = f(x)$ where $f$ is nonlinear. The Koopman operator $K$ acts on observables $g\colon X \to \mathbb{R}$ via $(Kg)(x) = g(\varphi^t(x))$, where $\varphi^t$ is the flow map. For any finite-dimensional Koopman-invariant subspace spanned by observables $g_1, \ldots, g_r$, the dynamics reduce to $z(t) = \exp(L \cdot t)\, z_0$ where $z = (g_1(x), \ldots, g_r(x))$ and $L \in \mathbb{R}^{r \times r}$ is the generator of the Koopman semigroup restricted to this subspace. The solution operator $K(t) = \exp(L \cdot t)$ and the generator $L$ encode the same information, but $L$ is \emph{diagnostic} while $K$ is not. The eigenvalues of $L$ directly give decay rates (real part) and oscillation frequencies (imaginary part). The matrix structure of $L$ enables interrogation of the underlying system physics. The generator of the lifted dynamics is linear although the lift is not.

\subsection{Architecture}

\paragraph{Encoder: nonlinear Koopman lift.}
The input consists of $T$ temporal snapshots of the $n \times n$ spatial field $\omega(x, y, T)$ concatenated with two coordinate channels, giving a per-pixel feature vector of dimension $T + 2$. A pointwise linear layer projects each pixel from $T + 2$ to $w$ dimensions. Two spectral convolution layers, each followed by a pointwise convolution and GELU nonlinearity, perform channel mixing in both Fourier space (per-mode, via learned complex weights at $M$ retained modes per dimension) and physical space (per-pixel, via $1 \times 1$ convolution). A final pointwise projection maps from $w$ to $r$ channels. The output $Z_0 \in \mathbb{R}^{r \times n \times n}$ is the Koopman observable field: $r$ latent channels at each of $n^2$ spatial points. The channel dimension $r$ is a hypothesis about the intrinsic dimensionality of the latent space; its effect on accuracy and physics recovery is studied in Appendix~\ref{app:channel}.

\paragraph{Generator: structured linear dynamics.}
The latent field $Z_0$ is transformed to Fourier space via FFT and truncated to $M$ modes per dimension, yielding $C_0 \in \mathbb{C}^{r \times M \times M}$. At each retained Fourier mode $k$, the dynamics are governed by an $r \times r$ generator matrix:
\begin{equation*}
L_k = S - D_k
\end{equation*}
where:
\begin{align*}
S &= (P - P^\top)/2, \quad P \in \mathbb{R}^{r \times r}, \\
D_k &= \mathrm{diag}\!\left( \mathrm{softplus}(d) + \mathrm{softplus}(\alpha) \cdot |k|^2 / k_\mathrm{max}^2 \right), \quad d, \alpha \in \mathbb{R}^r.
\end{align*}

$S$ is skew-symmetric by construction. It has $r(r-1)/2$ free parameters and is shared across all $M^2$ Fourier modes. This encodes the assumption that the conservative modal coupling structure is the same at every spatial frequency.

$D_k$ is diagonal and positive by construction (via softplus). It has $2r$ parameters: a base damping $d \in \mathbb{R}^r$ and a Laplacian coefficient $\alpha \in \mathbb{R}^r$. The dissipation at channel $i$ and mode $k$ is $d_i + \alpha_i |k|^2 / k_\mathrm{max}^2$. At $k = 0$, dissipation is $d_0$ alone. As $|k|^2$ grows, the $\alpha$ term dominates. This is the structure of viscous dissipation $\nu \nabla^2 \omega$ in Fourier space, where the Laplacian becomes multiplication by $-|k|^2$. The total number of interpretable parameters in the generator is $r(r-1)/2 + 2r$: the coupling structure plus the dissipation structure. These parameters govern the dynamics at all $M^2$ Fourier modes simultaneously.

\begin{proposition}[Architectural stability]\label{prop:stab}
Let $L_k = S - D_k$ where $S = (P - P^\top)/2 \in \mathbb{R}^{r \times r}$ and $D_k = \mathrm{diag}(\mathrm{softplus}(d) + \mathrm{softplus}(\alpha) \cdot |k|^2 / k_\mathrm{max}^2) \succeq 0$. Then every eigenvalue $\lambda$ of $L_k$ satisfies $\mathrm{Re}(\lambda) \le 0$.
\end{proposition}

\begin{proof}
For eigenvector $v \in \mathbb{C}^r$ with eigenvalue $\lambda$: $\lambda \|v\|^2 = v^* L_k v = v^* S v - v^* D_k v$. Since $S$ is real and skew-symmetric, $v^* S v$ is purely imaginary, so $\mathrm{Re}(v^* S v) = 0$. Since $D_k$ is real and positive-semidefinite (diagonal with non-negative entries), $v^* D_k v \ge 0$. Hence \mbox{$\mathrm{Re}(\lambda) \|v\|^2 = -v^* D_k v \le 0$}.
\end{proof}

\paragraph{Propagation: matrix exponential.}
The time-evolved Fourier coefficient at mode $k$ is:
\begin{equation*}
C_t[k] = \exp(L_k \cdot t) \cdot C_0[k], \quad \text{for any real } t \ge 0.
\end{equation*}
Since $L_k$ is $r \times r$ (typically small), the matrix exponential is computed via scaling-and-squaring approximation at negligible cost. The key property: $t$ is a continuous scalar. Changing $t$ from 1 to 200 does not require 200 sequential applications, it requires evaluating the same matrix exponential with a different scalar multiplier. The computational cost is $O(1)$ in the time horizon, versus $O(T)$ for autoregressive methods.

\paragraph{Decoder.}
The propagated coefficients replace the low-mode block of the initial Fourier representation; high-frequency modes beyond the retained $M \times M$ are carried forward from the initial encoding unchanged. An inverse FFT recovers $Z_t \in \mathbb{R}^{r \times n \times n}$. A pointwise two-layer network ($r \to \text{hidden} \to 1$) maps the $r$ latent channels at each pixel to a scalar prediction. The decoder is deliberately simple (pointwise, no spatial mixing) so that all spatial structure is forced through the generator.

\subsection{Training}
The model is trained end-to-end by minimizing the per-sample relative $L_2$ error between predicted and true trajectories over $T_\mathrm{out}$ timesteps. The $|k|^2$ form of $D_k$ is a modeling choice; the coefficients $d$, $\alpha$ and the matrix $S$ are learned from the prediction loss alone.

\section{Physics from the generator}

We evaluate LGN-KM on 2D Navier--Stokes turbulence at two viscosities: $\nu = 10^{-5}$ ($\mathrm{Re} \approx 10^5$, 1200 trajectories, $64 \times 64$, $T = 20$) and $\nu = 10^{-3}$ ($\mathrm{Re} \approx 10^3$, 5000 trajectories, $64 \times 64$, $T = 50$). Models use $T = 10$ input frames. Architecture: $r = 32$ channels, $M = 12$ Fourier modes, width $w = 32$. Training: 500 epochs, AdamW optimizer, cosine scheduling. FNO-2D with matched mode count serves as the baseline. Full details in Appendix~\ref{app:training}. The generator at each viscosity has 560 interpretable parameters: 496 in $S$ (coupling) and 64 in $D$ (32 base damping values $d$ and 32 Laplacian coefficients $\alpha$) which are extracted directly from the learned weights with no post-hoc fitting.

\subsection{Eigenspectra of nonlinear PDE}

\begin{figure}[!t]
\centering
\begin{subfigure}{0.32\linewidth}
  \centering
  \includegraphics[width=\linewidth]{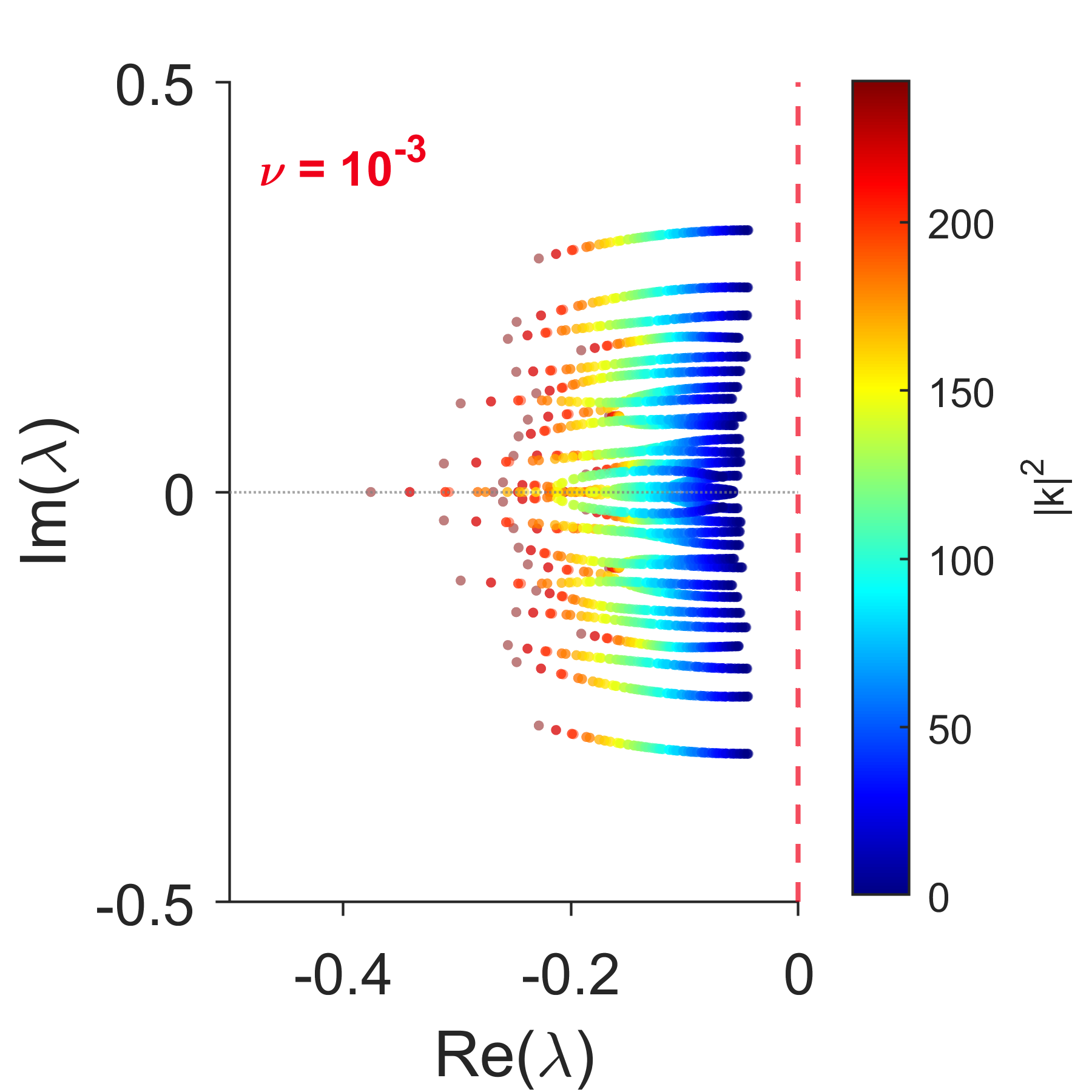}
  \caption{}
\end{subfigure}
\hfill
\begin{subfigure}{0.32\linewidth}
  \centering
  \includegraphics[width=\linewidth]{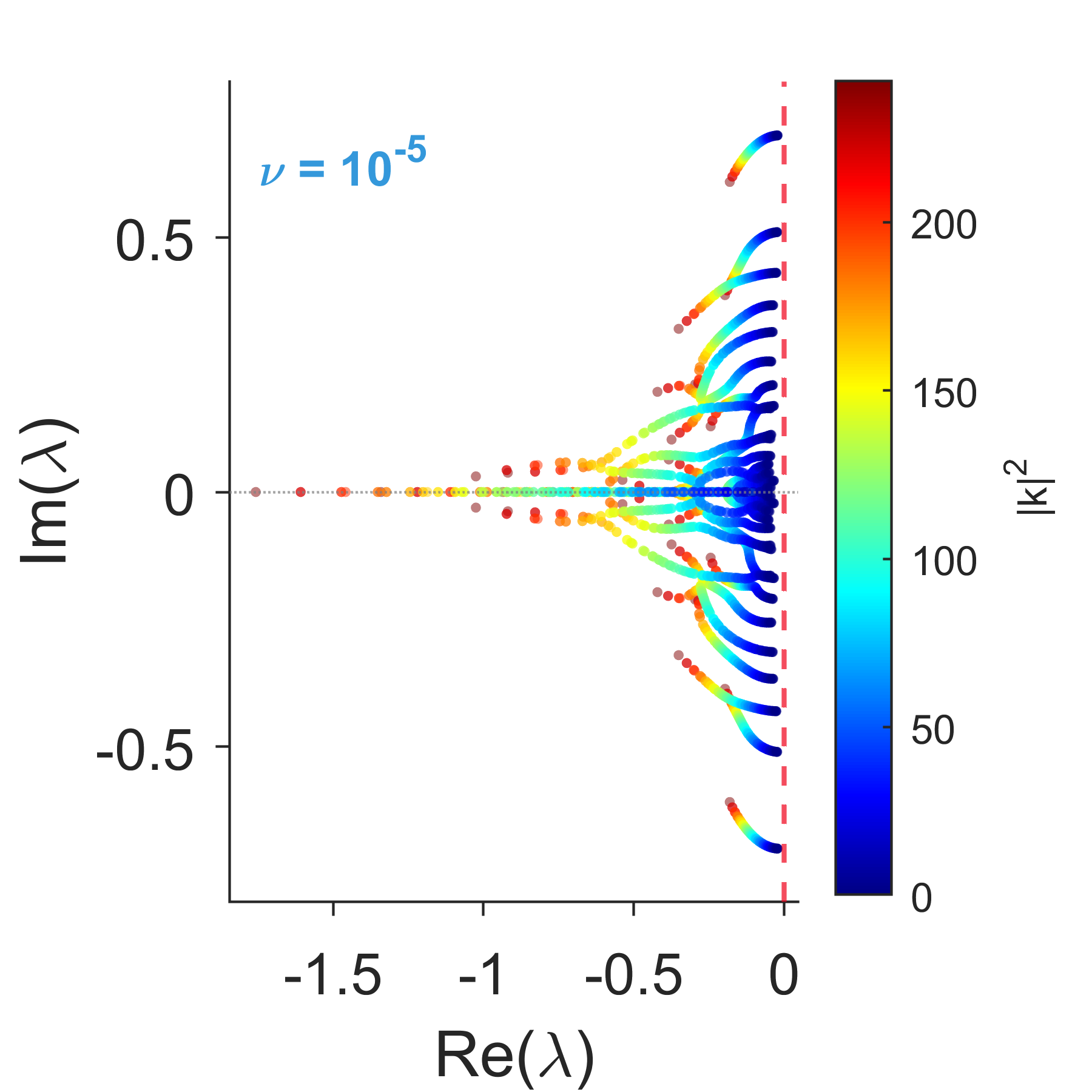}
  \caption{}
\end{subfigure}
\hfill
\begin{subfigure}{0.32\linewidth}
  \centering
  \includegraphics[width=\linewidth]{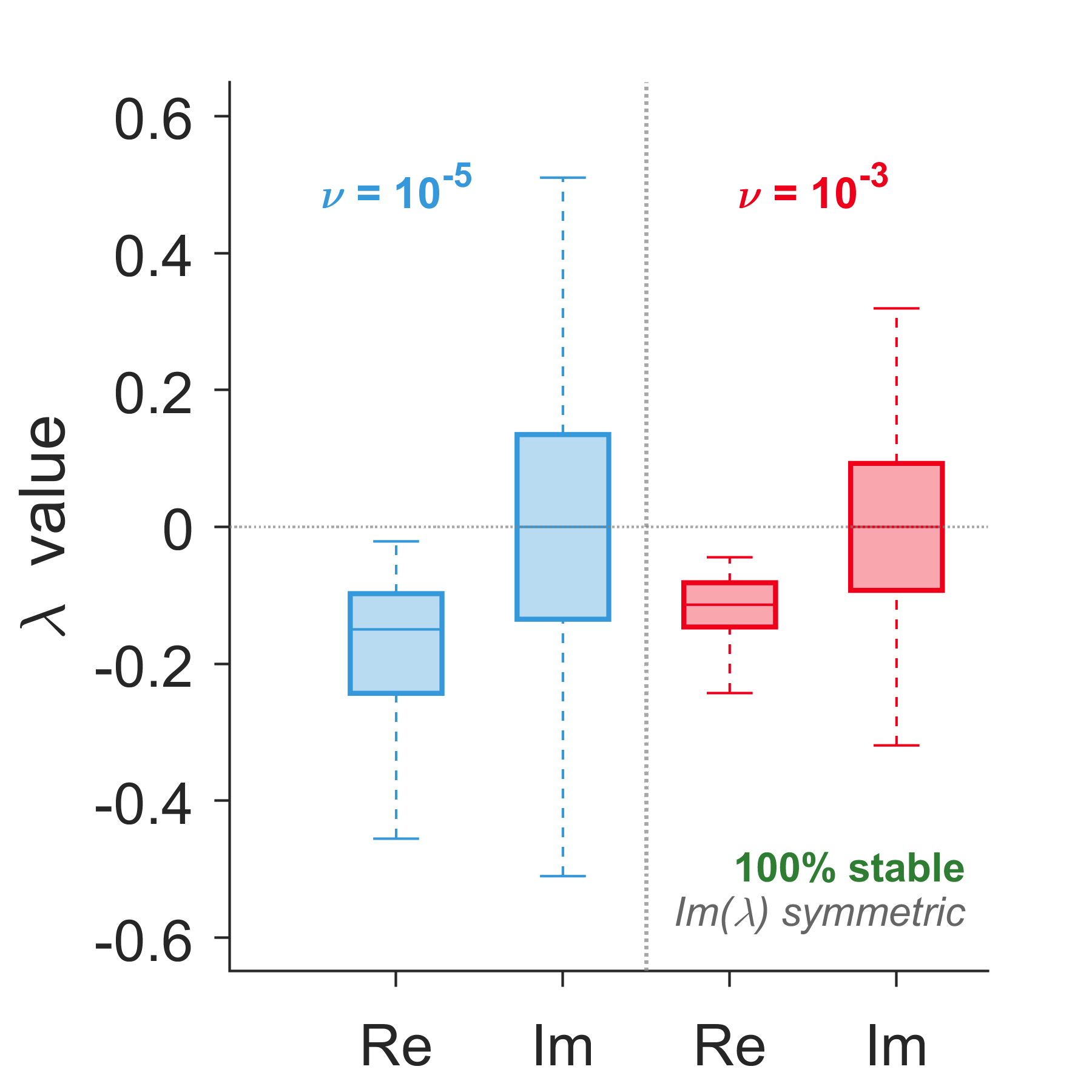}
  \caption{}
\end{subfigure}
\caption{Eigenspectra of the learned Koopman generator. (a,b) All 4{,}608 eigenvalues of the Koopman generator on the complex plane, colored by wavenumber $|k|^2$. Each of the $r$ ($32$) dispersion branches traces a continuous curve from the imaginary axis into the left half-plane and divides into conjugate pairs from skew-symmetric $S$. (c) $\mathrm{Re}(\lambda) < 0$ by construction shifts between regimes and $\mathrm{Im}(\lambda)$ remains symmetric about zero. At $\nu = 10^{-3}$ the spectrum is compact relative to $\nu = 10^{-5}$.}
\label{fig:eigenspectra}
\end{figure}

Nonlinear PDEs such as Navier--Stokes equations have no eigenspectrum. The quadratic nonlinearity $(u \cdot \nabla)\omega$ breaks superposition; linearization around a base flow yields local stability modes, not global spectral structure. Here we show that the Koopman generator of 2D Navier--Stokes turbulence has a complete, stable, and physically interpretable eigenspectrum, learned from trajectory prediction. The generator $L_k$ at each Fourier mode $k$ is an $r \times r$ matrix with $r$ eigenvalues. Across all $M^2$ retained modes, this yields $r \times M^2 = 4{,}608$ eigenvalues. Figure~\ref{fig:eigenspectra} plots the full set on the complex plane, colored by wavenumber magnitude $|k|^2$.

The figure reveals $r$ ($32$) dispersion branches $\lambda_r(k)$, coherent curves traced by each eigenvalue as $|k|^2$ increases from $0$ to $242$. The $S{-}D_k$ decomposition is directly legible in the geometry. The vertical extent of each branch (oscillation frequencies) is bounded by the largest singular value of $S$ and shaped jointly by $S$ and $D_k$; it remains bounded regardless of wavenumber. The horizontal extent (decay rates) is governed by $D_k$ which grows linearly with $|k|^2$. Smaller wavenumber modes cluster near the imaginary axis where dynamics are oscillatory and weakly damped whereas high frequency modes extend far into the left half-plane where viscous dissipation dominates. The $16$ branches in the upper half-plane mirror those below as they are conjugate pairs guaranteed by $S$ being real and skew-symmetric. These are dispersion relations in the classical sense; the same object that characterizes phonons in crystals, electrons in semiconductors, and photons in waveguides, here extracted from a neural network trained on turbulence data.

At $\nu = 10^{-3}$ (Fig.~\ref{fig:eigenspectra}a), the $32$ branches are short and tightly bundled: $\mathrm{Re}(\lambda) \in [-0.38, -0.04]$, $\mathrm{Im}(\lambda) \in [-0.32, 0.32]$. High viscosity suppresses fine-scale dynamics before they develop separated timescales, compressing the spectrum into a narrow region of the complex plane. At $\nu = 10^{-5}$ (Fig.~\ref{fig:eigenspectra}b), the same $32$ branches unfold: $\mathrm{Re}(\lambda)$ extends to $-1.76$ and $\mathrm{Im}(\lambda)$ spans $\pm 0.70$; a $5.2\times$ wider range in decay rates and $2.2\times$ in oscillation frequencies. The Frobenius norm of $S$ is $1.82\times \pm 0.14$ larger at $\nu = 10^{-5}$ than at $\nu = 10^{-3}$; stronger advective coupling produces the $2.2\times$ vertical expansion directly. The box plots (Fig.~\ref{fig:eigenspectra}c) confirm: $\mathrm{Re}(\lambda)$ distributions shift substantially between regimes while $\mathrm{Im}(\lambda)$ remains symmetric about zero at both viscosities, consistent with conjugate symmetry of $S$. Every eigenvalue across both models satisfies $\mathrm{Re}(\lambda) < 0$, with $\max \mathrm{Re}(\lambda) = -0.0211$, constrained architecturally. The spectra in Figure~\ref{fig:eigenspectra} are, to our knowledge, the first complete structurally stable dispersion relation extracted from a neural operator trained on turbulent flow. Structural ablations confirm this requires the full $S{-}D$ decomposition; an unconstrained generator produces dispersion branches but $41\%$ of eigenvalues are unstable, $S$-only collapses to the imaginary axis, and $D$-only to the real axis (Appendix~\ref{app:ablation}, Table~\ref{tab:ablation}).

\subsection{Physics recovery and universality}
\begin{figure}[!t]
\centering
\begin{subfigure}{0.24\linewidth}
  \centering
  \includegraphics[width=\linewidth]{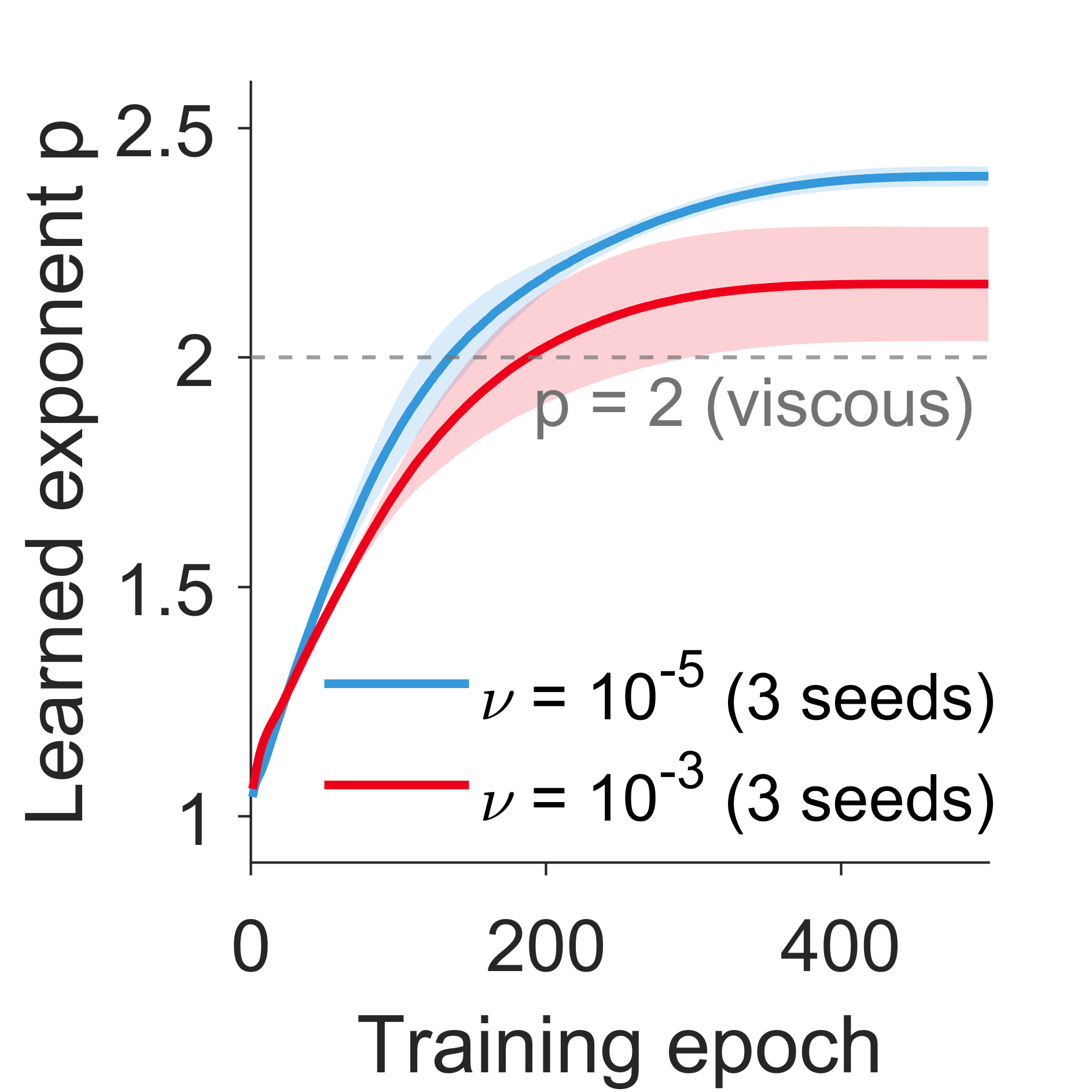}
  \caption{}
\end{subfigure}
\hfill
\begin{subfigure}{0.24\linewidth}
  \centering
  \includegraphics[width=\linewidth]{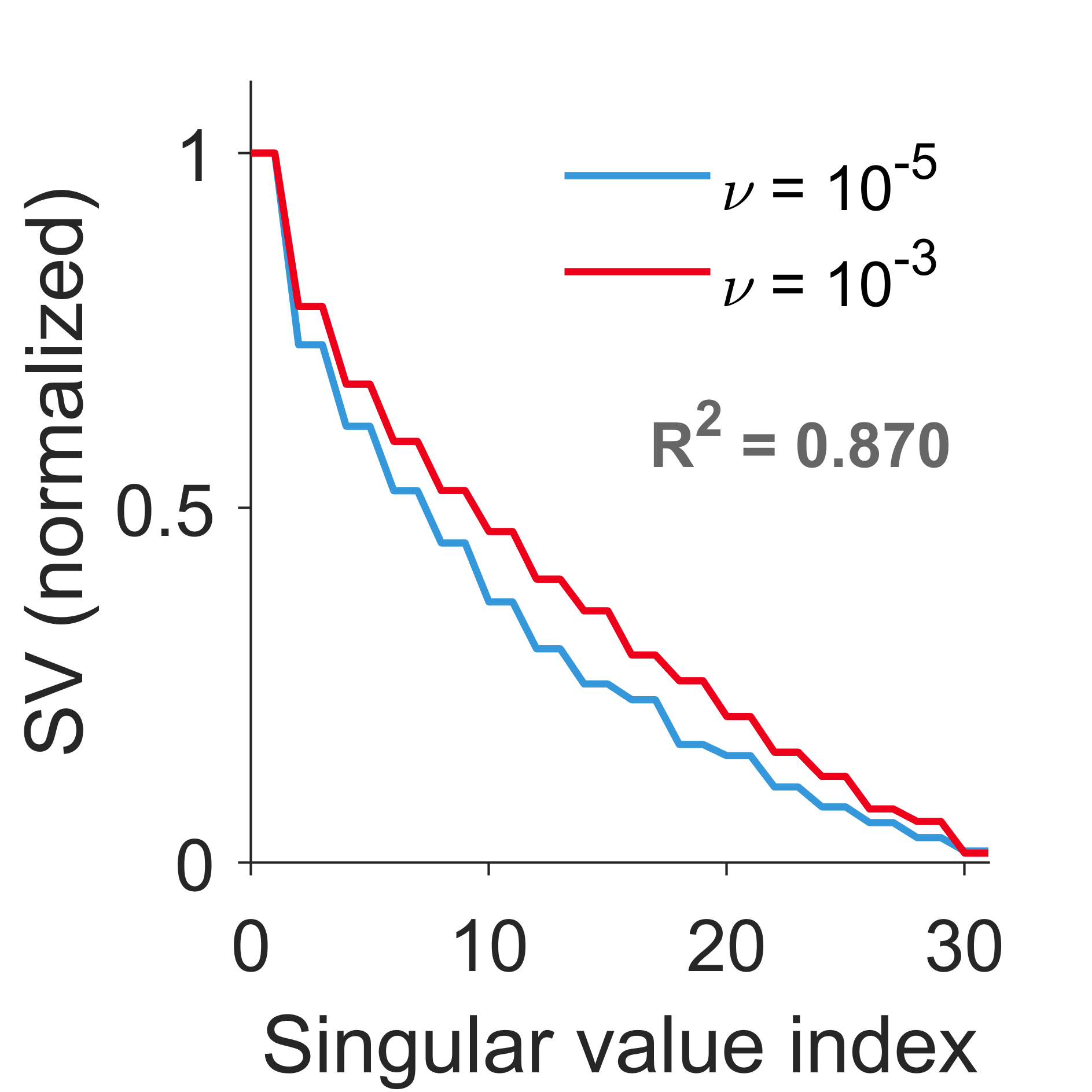}
  \caption{}
\end{subfigure}
\hfill
\begin{subfigure}{0.24\linewidth}
  \centering
  \includegraphics[width=\linewidth]{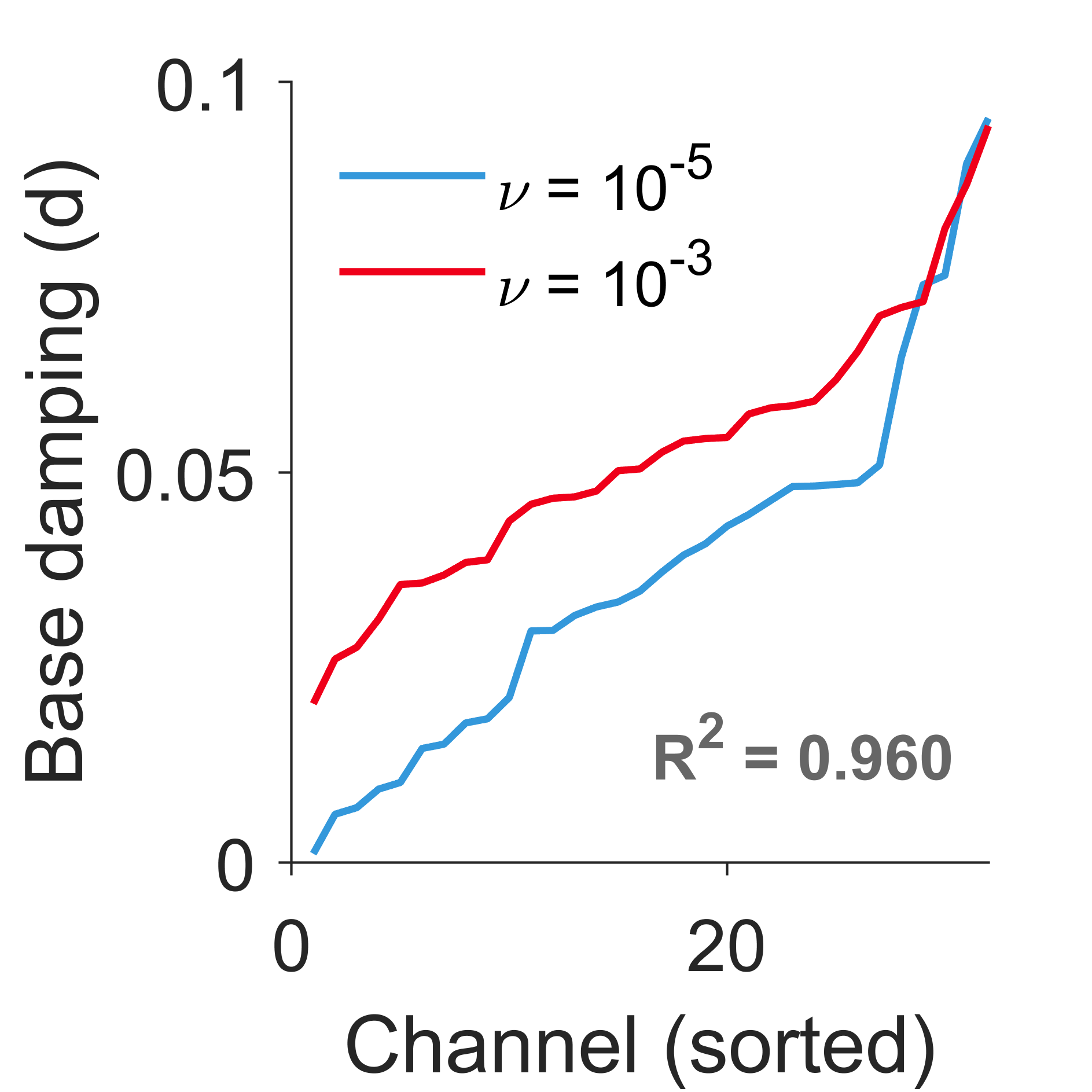}
  \caption{}
\end{subfigure}
\hfill
\begin{subfigure}{0.24\linewidth}
  \centering
  \includegraphics[width=\linewidth]{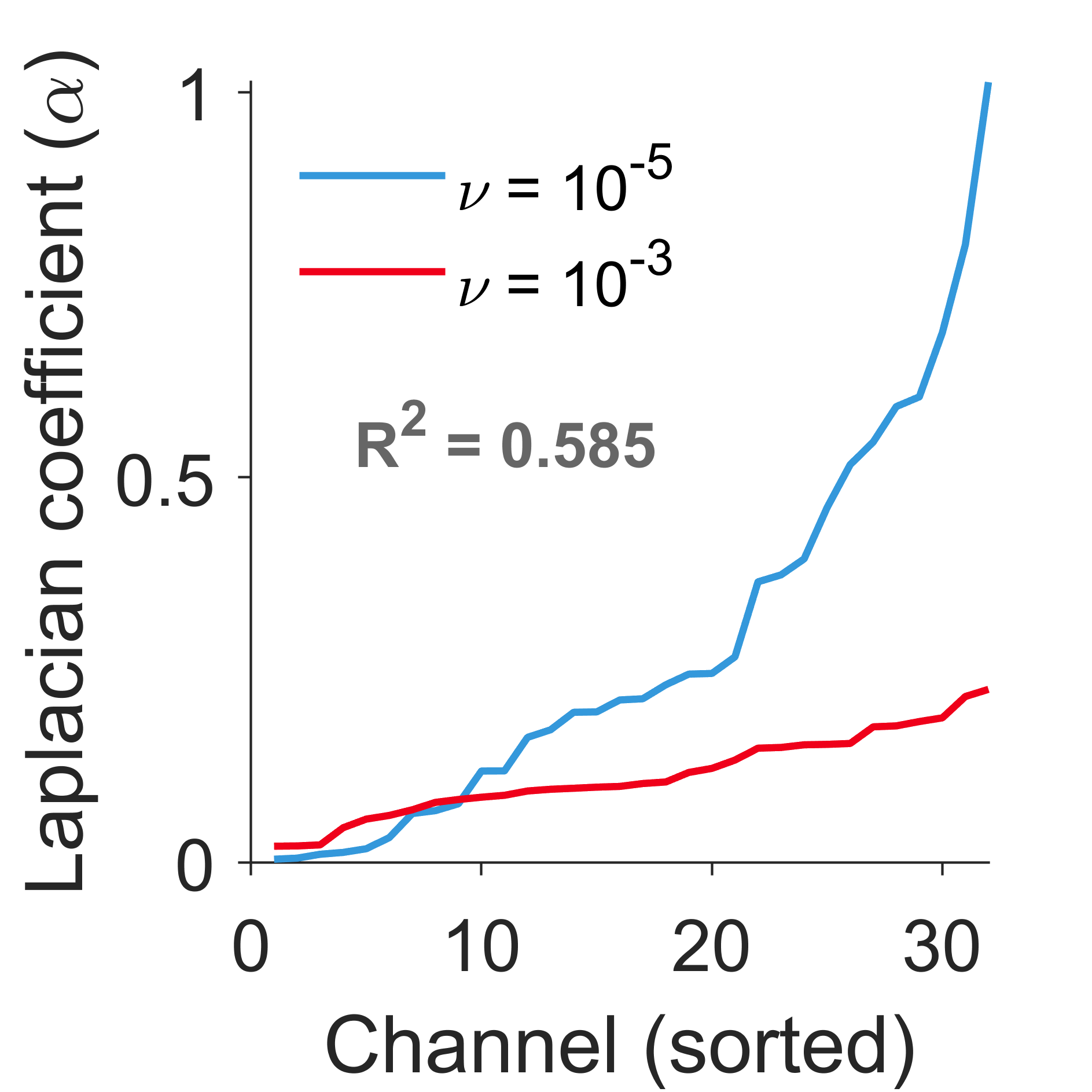}
  \caption{}
\end{subfigure}
\caption{Physics from the learned generator on 2D Navier--Stokes. (a) Learned exponent $p$ under $D_k = \mathrm{diag}\!\left(\mathrm{softplus}(d) + \mathrm{softplus}(\alpha)\cdot |k|^{p}/k_{\max}^{p}\right)$ initialized at $p = 1$. (b--d) Models with fixed $p = 2$. (b) Normalized SVD profiles of $S$. (c) Sorted base damping distribution match across viscosities. (d) Sorted Laplacian coefficient $\alpha$. All quantities extracted directly from learned parameters. $R^{2}$ is computed on $[0,1]$-normalized profiles to compare shape independent of scale.}
\label{fig:universality}
\end{figure}

The main model prescribes quadratic wavenumber dependence in $D_k$ and learns the damping coefficients $d$ and $\alpha$. To test whether the model can directly learn it from data, we replace $|k|^2$ by $|k|^p$ with $p$ a learnable scalar initialized at $p = 1$ and train under the same protocol. Driven by the prediction loss alone, $p$ rises through 2 and ends at $2.39 \pm 0.02$ ($\nu = 10^{-5}$) and $2.16 \pm 0.12$ ($\nu = 10^{-3}$) over three seeds (Fig.~\ref{fig:universality}a), with test error unchanged ($0.241$ vs $0.240$; $0.062$ vs $0.063$). These results support approximately quadratic damping within the tested power-law family. Removing the functional form entirely leaves accuracy unchanged but makes the fitted exponent range-dependent (Appendix~\ref{app:exponent}); the spectral readout is insensitive to $r$, $M$ and input normalization (Appendix~\ref{app:sensitivity}). The fixed $p = 2$ models of Section~3.1 learn in different latent coordinates, which raises the question of what can be compared between them.

\begin{proposition}[Gauge invariance of the singular spectrum]\label{prop:gauge}
Let $\Phi \in O(r)$ be an orthogonal change of latent coordinates and $S' = \Phi S \Phi^\top$ the transformed coupling matrix. Then $\sigma(S') = \sigma(S)$; the element-wise cosine similarity $\langle \mathrm{vec}(S), \mathrm{vec}(S') \rangle / (\|S\|_F \|S'\|_F)$ depends on $\Phi$.
\end{proposition}

\begin{proof}
Singular values are invariant under orthogonal conjugation. Element-wise cosine similarity equals $1$ iff $\Phi S \Phi^\top = S$, and for generic $\Phi \in O(r)$ outside this stabilizer it takes values in $[-1, 1]$ depending on $\Phi$.
\end{proof}

\begin{figure}[!b]
\centering
\begin{subfigure}{0.24\linewidth}
  \centering
  \includegraphics[width=\linewidth]{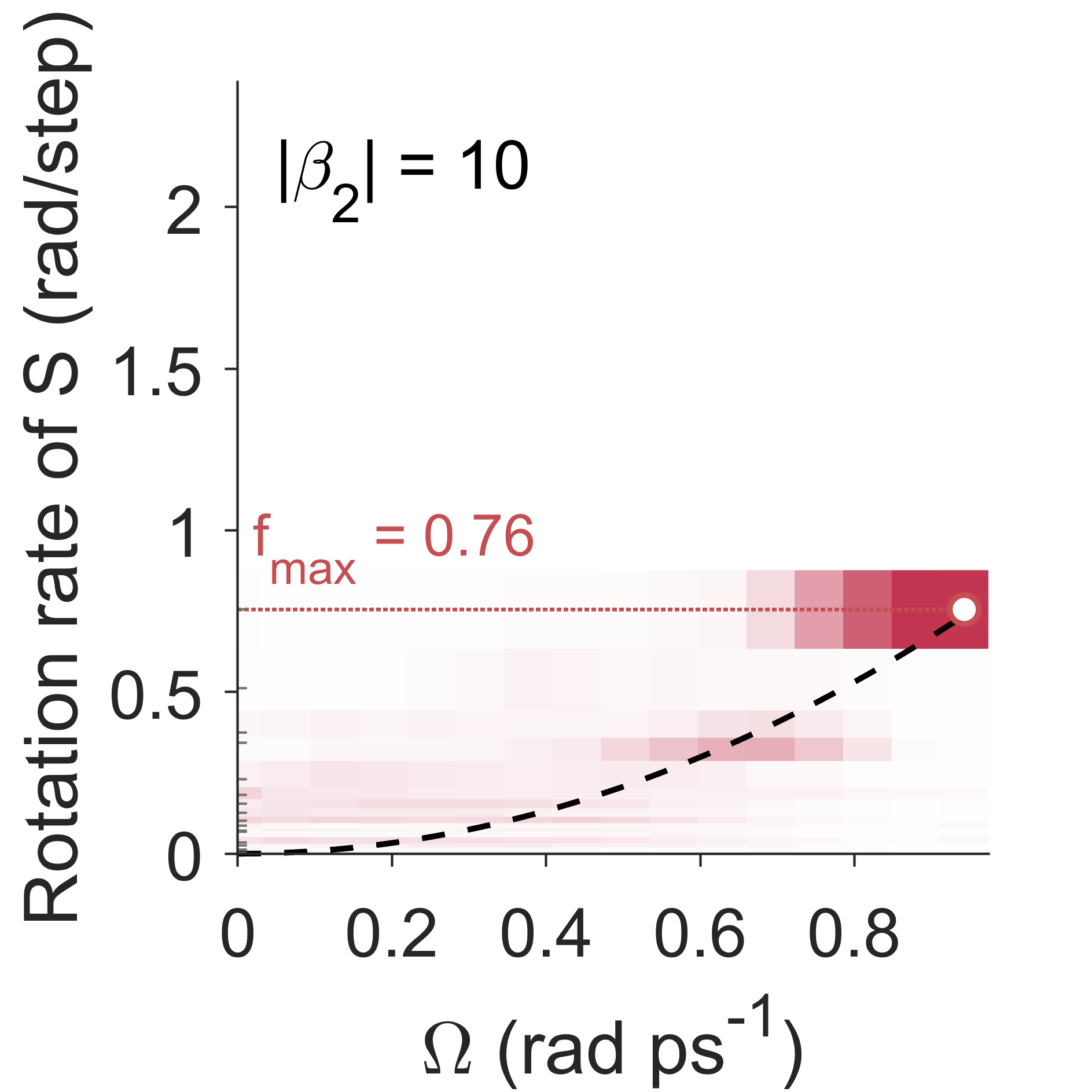}
  \caption{}
\end{subfigure}
\hfill
\begin{subfigure}{0.24\linewidth}
  \centering
  \includegraphics[width=\linewidth]{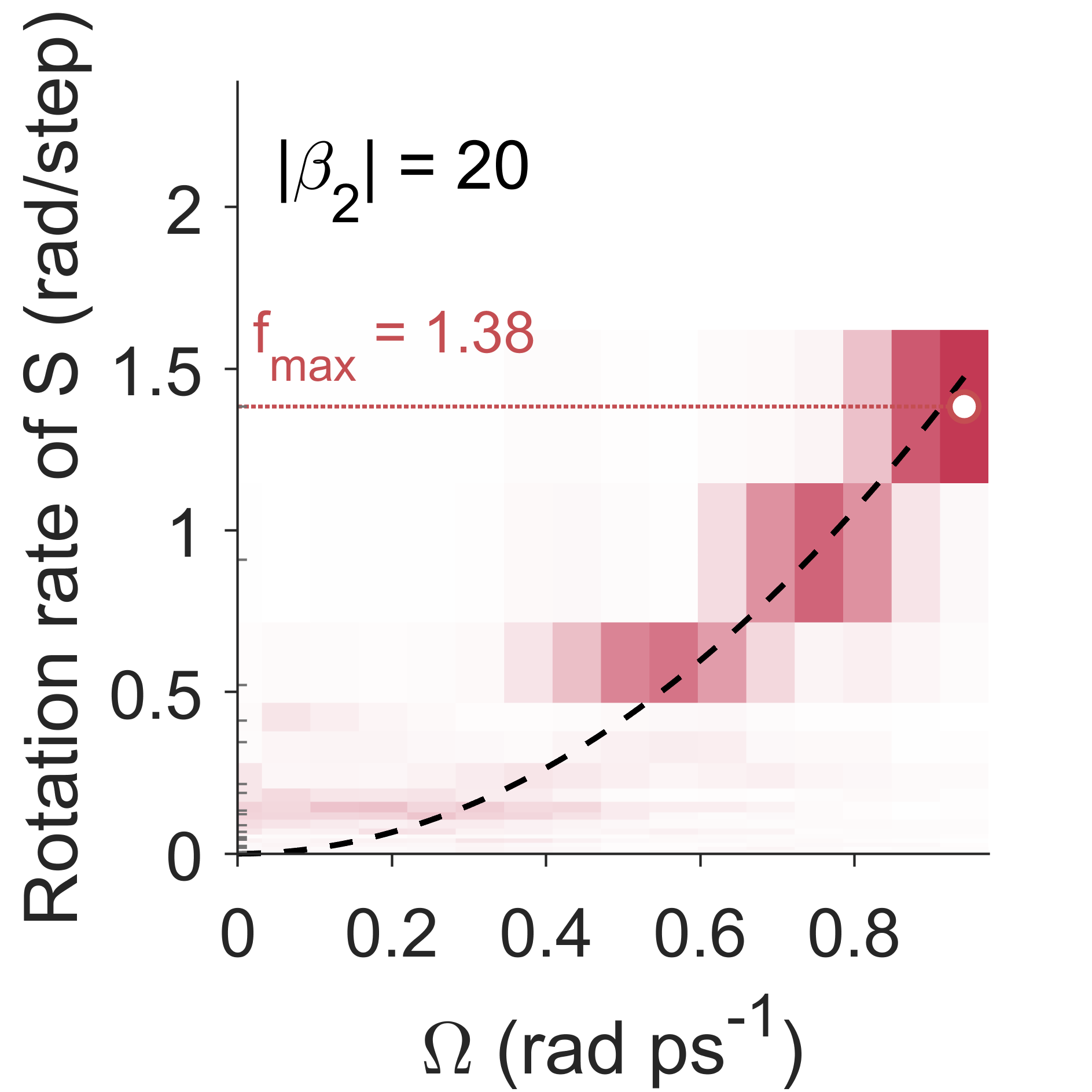}
  \caption{}
\end{subfigure}
\hfill
\begin{subfigure}{0.24\linewidth}
  \centering
  \includegraphics[width=\linewidth]{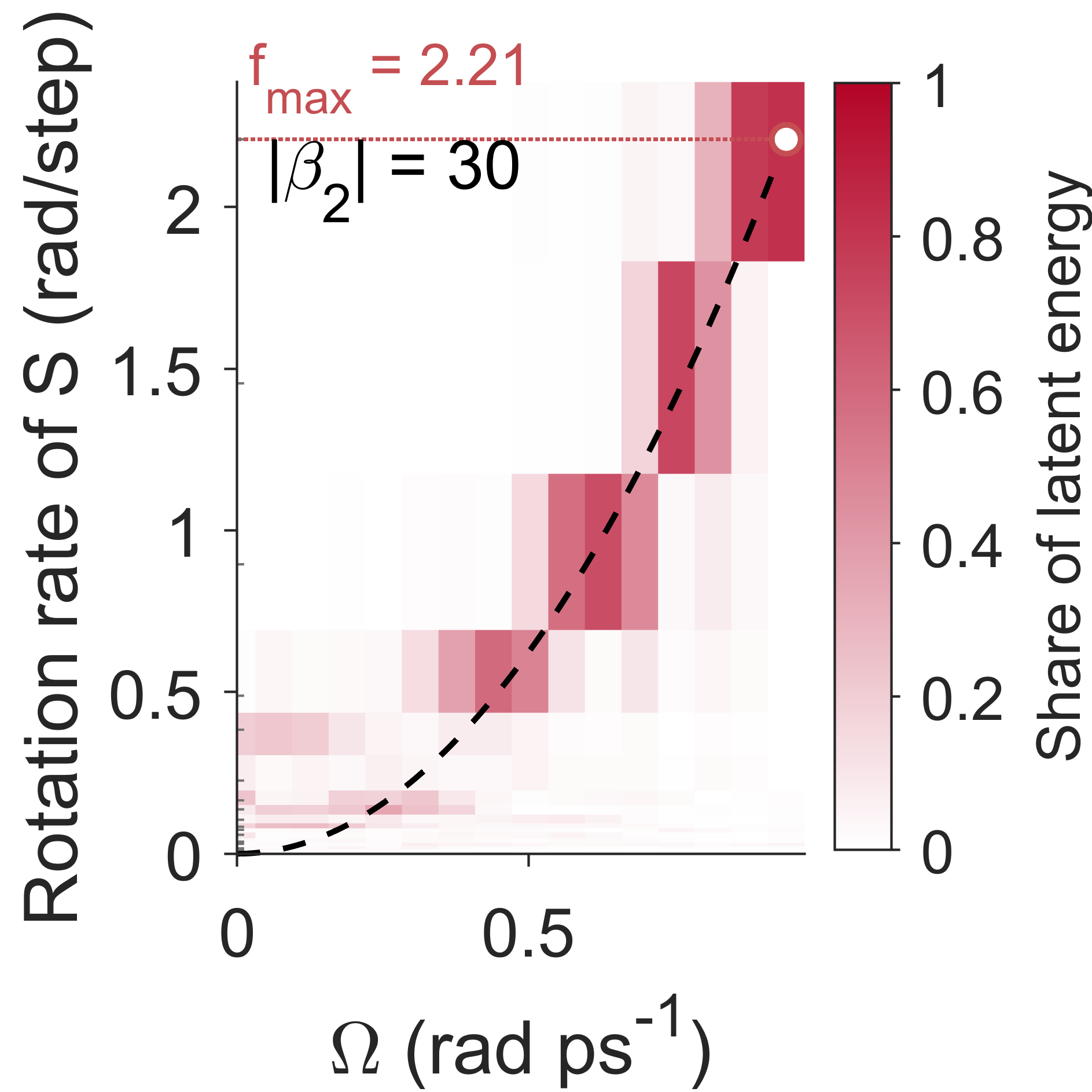}
  \caption{}
\end{subfigure}
\hfill
\begin{subfigure}{0.24\linewidth}
  \centering
  \includegraphics[width=\linewidth]{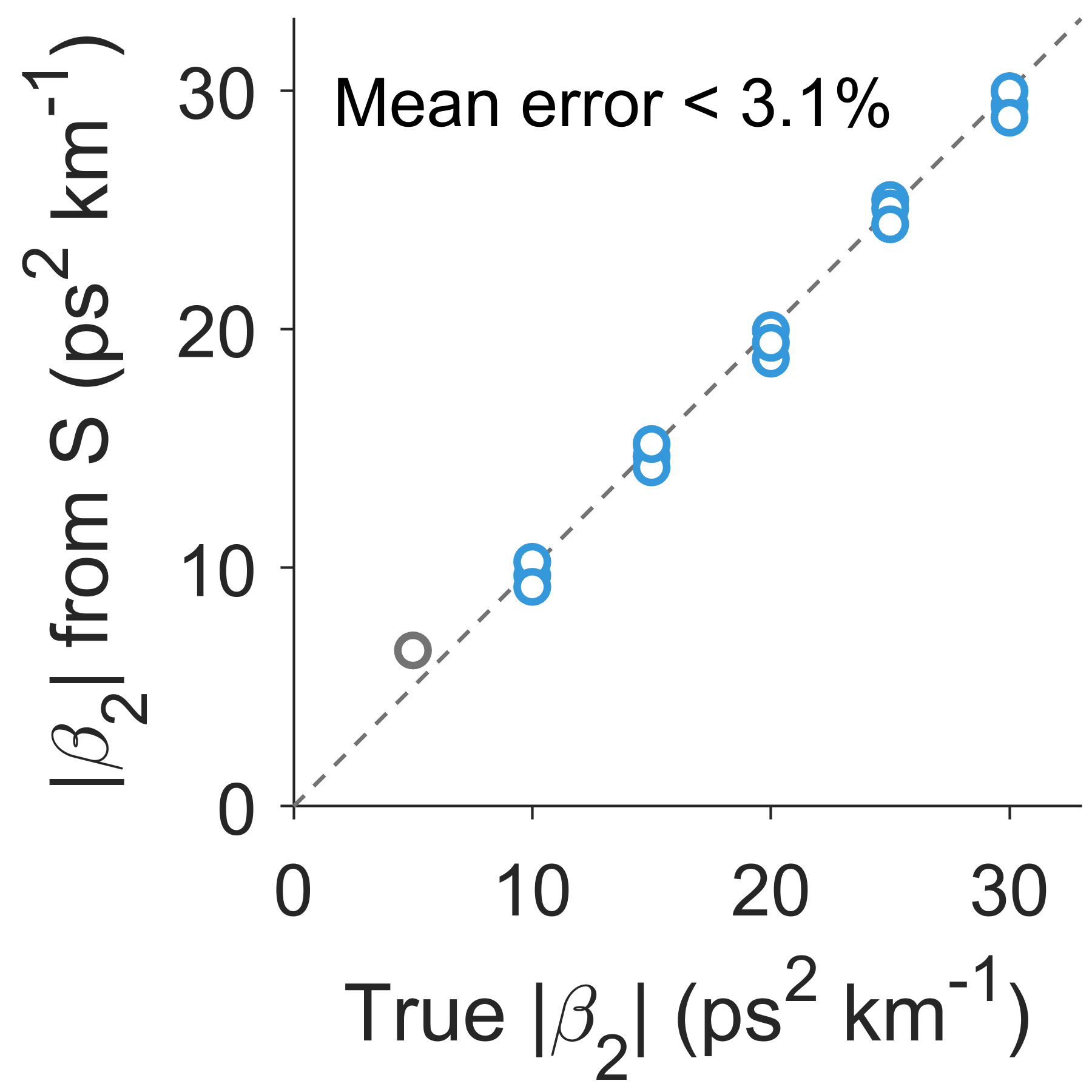}
  \caption{}
\end{subfigure}
\caption{The rotation rates of $S$ track the fiber's dispersion relation. (a--c) For $|\beta_2| = 10$, $20$ and $30$~ps$^2$\,km$^{-1}$ (seed 1), the share of each input frequency $\Omega$'s latent energy carried by each rotation rate of $S$ on held-out pulses. Dashed lines show the predicted phase advance from the true $\beta_2$. The fastest rotation $f_{\max}$ sits at the band edge $\Omega_{15}$. (d) Predicted vs true $|\beta_2|$ for five different fibers; the three-seed mean is within $3.1\%$ of the true value at every fiber, seed standard deviation $1.9$--$5.3\%$.}
\label{fig:nlse}
\end{figure}

Therefore, their raw $S$ matrices have cosine similarity $0.067 \pm 0.058$, reflecting a gauge symmetry of the Koopman lifting, although the gauge-invariant quantities are consistent across two flow regimes. The singular value profiles of $S$ match at $R^2 = 0.870 \pm 0.063$ (Fig.~\ref{fig:universality}b), which shows that the two models learn similar normalized coupling spectra across the tested regimes, although the norms of $S$ differ (Section~3.1). Similarly, the sorted $d$ distributions match across viscosities at $R^2 = 0.960 \pm 0.017$. Base damping $d$ captures the dissipation at $k = 0$.

Since the viscous term $\nu |k|^2$ vanishes at $k = 0$, $d$ reflects intrinsic decay timescales of the observables independent of viscosity. The Laplacian coefficient $\alpha$, which governs wavenumber-dependent dissipation, does not match across viscosities ($R^2 = 0.585 \pm 0.273$; wider variance due to regime-specific component) which is physically expected. The generator thus separates the conservative coupling $S$ and base damping $d$, from regime-specific dissipation governed by the Laplacian coefficient $\alpha$. The same decomposition generalizes beyond fluid mechanics; on 2D FitzHugh--Nagumo reaction-diffusion, the generator shows matched $S$ singular-value profiles ($R^2 = 0.994$, single seed) with no architecture modification (Appendix~\ref{app:fhn}).

\subsection{Testing the learned generator against known optical dispersion}

The Navier--Stokes results show how decay rates vary with wavenumber under the prescribed Laplacian damping. We now ask whether the rotation frequencies learned by $S$ track a known physical law. Pulse propagation in an optical fiber provides a direct test through the nonlinear Schr\"odinger equation:
\begin{equation*}
\frac{\partial E}{\partial z} = -\,i\frac{\beta_2}{2}\frac{\partial^2 E}{\partial T^2} + \frac{\beta_3}{6}\frac{\partial^3 E}{\partial T^3} + i\gamma |E|^2 E - \frac{\alpha}{2} E.
\end{equation*}
Here $E(z, T)$ is the complex field envelope, $z$ is propagation distance, and $T$ is time in a frame moving with the pulse. The coefficients $\beta_2$ and $\beta_3$ are group-velocity and third-order dispersion coefficients, $\gamma$ is the Kerr nonlinearity coefficient, and $\alpha$ is the fiber loss coefficient. Second-order dispersion preserves each Fourier mode's energy and advances its phase by a magnitude $|\beta_2|\Omega_k^2 \Delta z/2$ at angular frequency $\Omega_k$ over a propagation step $\Delta z$. We train LGN-KM on five simulated fibers with $|\beta_2|$ from 10 to 30~ps$^2$/km, three seeds per fiber, and $\Delta z = 0.166$~km. We use the same training process described in Section~2 with 16 Fourier modes, and a decoder that outputs the real and imaginary parts of $E$.

The same matrix $S$ acts at every optical frequency, so each Fourier mode has access to the same 16 rotation rates. The encoder determines how strongly each mode excites each rotation. On held-out pulses, the rotation carrying the most energy tracks the predicted quadratic phase advance (Fig.~\ref{fig:nlse}a--c). Along the band, the learned rotation rate is on average $0.96$--$1.19$ times the predicted one across all 15 models. At the highest retained frequency $\Omega_{15}$, the fastest learned rotation $f_{\max}$ (radians per snapshot) yields the dispersion estimate with no fitted constant, $|\hat{\beta}_2| = 2 f_{\max}/(\Omega_{15}^2 \Delta z) = 13.56\, f_{\max}$.

\begin{figure}[!t]
\begin{subfigure}{0.32\linewidth}
  \centering
  \includegraphics[height=4.5cm, keepaspectratio]{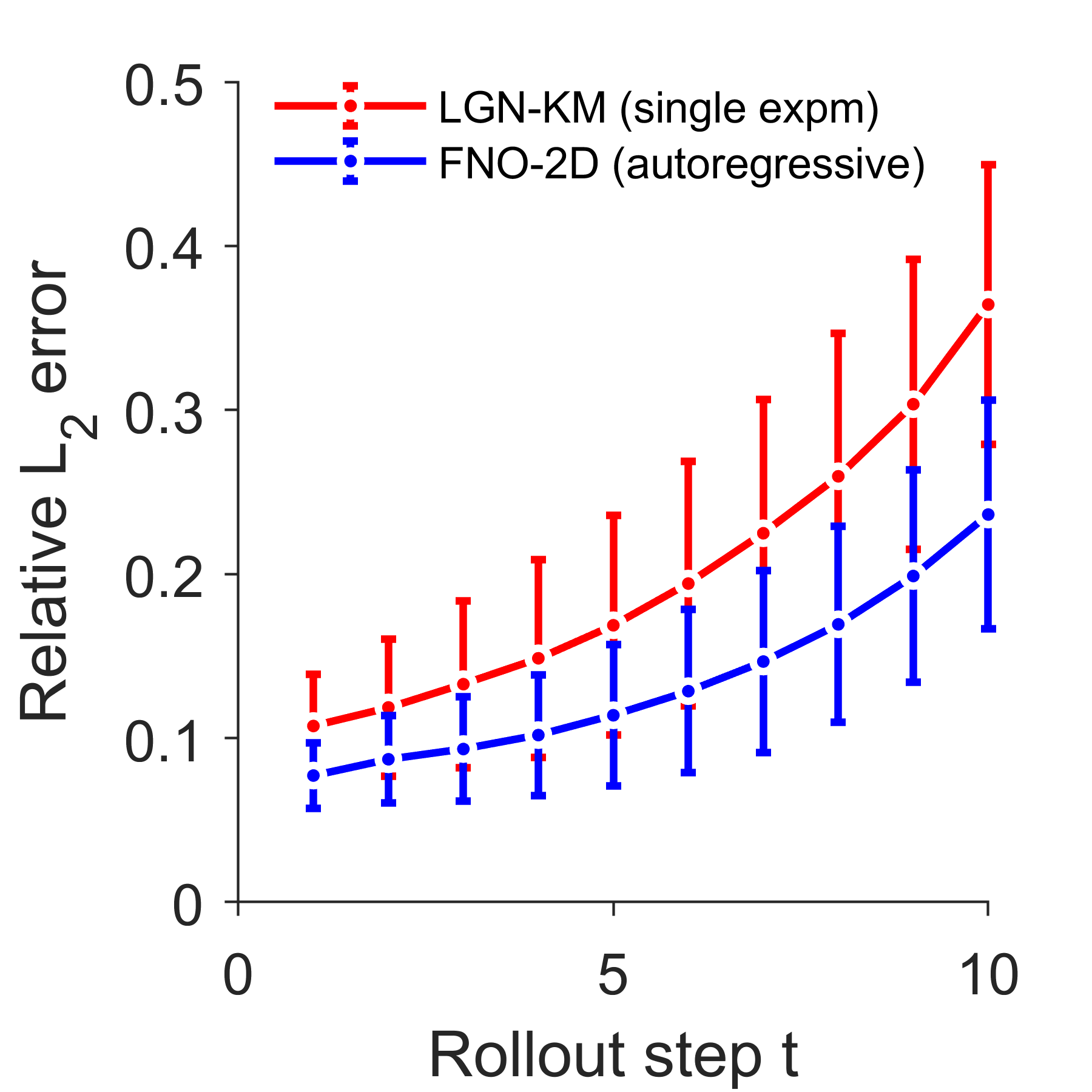}
  \caption{}
\end{subfigure}
\hfill
\begin{subfigure}{0.33\linewidth}
  \centering
  \includegraphics[height=4.5cm, keepaspectratio]{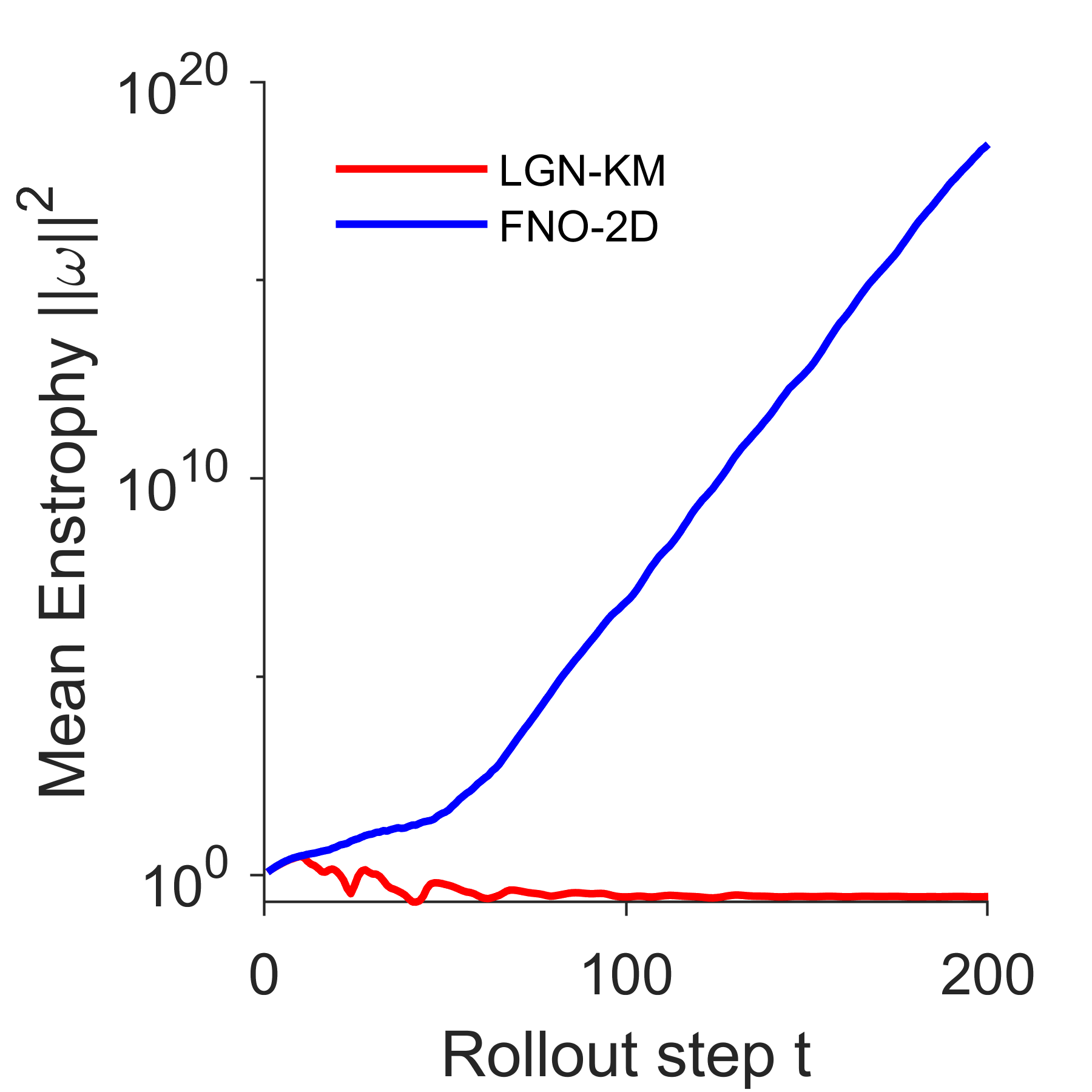}
  \caption{}
\end{subfigure}
\hfill
\begin{subfigure}{0.33\linewidth}
  \centering
  \includegraphics[height=4.5cm, keepaspectratio]{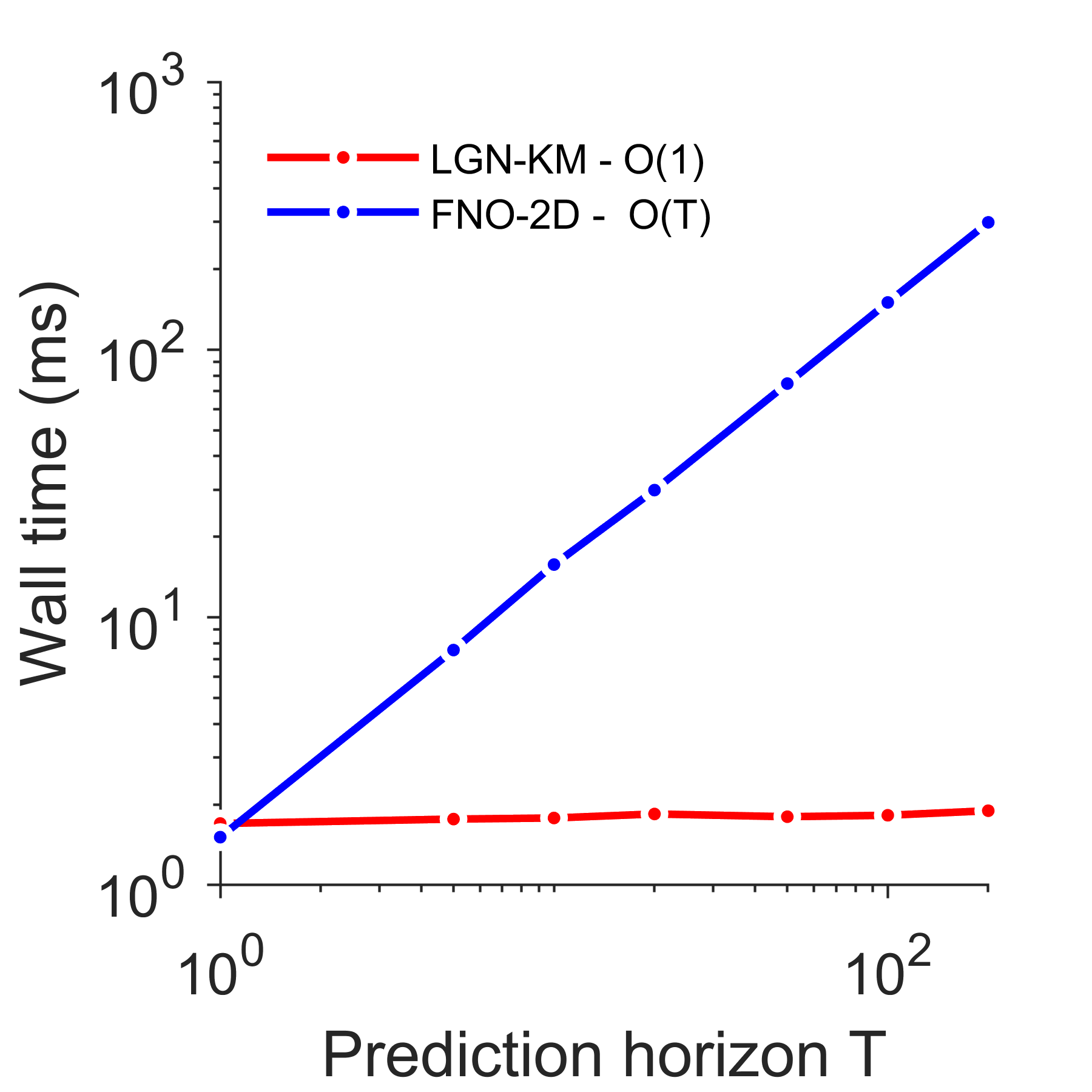}
  \caption{}
\end{subfigure}
\caption{Long-horizon stability. (a) Per-step relative $L_{2}$ error on held-out test trajectories within the training horizon ($T_{\mathrm{out}} = 10$); FNO-2D from our reimplementation of [1]. (b) Mean enstrophy $\|\omega\|^{2}$ over $200$ steps ($20\times$ beyond training horizon). FNO energy diverges to ${\sim}10^{18}$; LGN-KM stays bounded, guaranteed by the contractive eigenspectrum of the $S{-}D$ generator. (c) Wall time vs prediction horizon. LGN-KM evaluates any horizon via a single matrix exponential at $O(1)$ cost; FNO requires $T$ autoregressive steps.}
\label{fig:longhorizon}
\end{figure}

Across the five fibers, the three-seed mean is within $3.1\%$ of the true $|\beta_2|$ (Fig.~\ref{fig:nlse}d). The readout needs pulses that excite the band edge; at a true $|\beta_2|$ of 20~ps$^2$/km, wider pulses ($T_0 = 5$--$10$~ps) give $15.7$~ps$^2$/km and narrower pulses ($T_0 = 2$--$4$~ps) give $23.5$~ps$^2$/km (Appendix~\ref{app:nlse}). Neither the parameterization of $S$ nor the trajectory loss encodes the $\Omega^2$ phase law. The agreement between the learned rotation rates and the independently predicted phase advance therefore shows that the generator used to forecast the optical field also encodes its linear dispersion.

\section{Utilities of stable differentiable generator}

LGN-KM provides guaranteed stability, continuous-time evaluation at $O(1)$ cost, and physics-informed cross-viscosity model transfer.

\subsection{Long-horizon stability and \texorpdfstring{$O(1)$}{O(1)} continuous-time evaluation}

Since the symmetric part of $L_k$ is $-D_k \preceq 0$, $\frac{d}{dt}\|C_t[k]\|^2 = -2\, C_t[k]^* D_k C_t[k] \le 0$, so $\exp(L_k \cdot t)$ is contractive on the latent state for all $t \ge 0$; latent energy cannot grow regardless of rollout horizon. This is contractivity of the latent norm, not monotone decay of the decoded field. FNO has no analogous constraint. Each autoregressive step applies a learned nonlinear map with no contractivity guarantee, and small per-step errors compound multiplicatively. Figure~\ref{fig:longhorizon}(a) shows that FNO is more accurate than LGN-KM at every step within the training horizon, consistent with the accuracy trade-off discussed in Section~\ref{sec:disc}. LGN-KM energy stays bounded, consistent with the contractive eigenspectrum (Fig.~\ref{fig:longhorizon}b). FNO energy grows exponentially, reaching ${\sim}10^{18}$ at $T = 200$.

Baseline comparisons under the identical protocol are in Table~\ref{tab:baselines} and long-horizon accuracy in Appendix~\ref{app:longhorizon}. Within the training horizon FNO and KNO are more accurate. Beyond $t \approx 30$ no model tracks the flow, whose enstrophy grows from $1.2$ to $53$ under the benchmark's forcing. LRAN leaves the physical manifold at $t = 37$ and FNO at $t = 81$, while LGN-KM and KNO remain bounded to $t = 200$, with LGN-KM's error crossing below FNO's at $t \approx 33$ and approaching $1.0$, the value of a prediction decayed to the mean. The contractive generator cannot sustain energy injection; an input term $\dot{z} = Lz + Bu$ is the natural extension.

Since $t$ enters $\exp(L_k \cdot t)$ as a continuous scalar, LGN-KM evaluates at any real-valued time via a single matrix exponentiation. The cost is $O(1)$ in the prediction horizon. Evaluating at $T = 200$ costs $1.9$~ms, comparable to $1.7$~ms at $T = 1$. FNO requires $T$ sequential autoregressive steps, scaling linearly to $299$~ms at $T = 200$ (Fig.~\ref{fig:longhorizon}c). The same property gives predictions at times never seen in training. On ground truth sampled four times finer than the training frames, LGN-KM gives mean relative $L_2$ $0.190$, better at every query time than linearly interpolating the model's own integer-time predictions ($0.200$; Appendix~\ref{app:longhorizon}).

\begin{table}[!t]
\caption{Test relative $L_2$ under the paper's protocol. DMD is the best of ranks 32--256.}
\label{tab:baselines}
\centering
\footnotesize
\setlength{\tabcolsep}{4pt}
\begin{tabular}{lccp{2.9cm}p{3.75cm}}
\toprule
Model & $\nu = 10^{-3}$ (40 steps) & $\nu = 10^{-5}$ (10 steps) & Generator & Stability \\
\midrule
DMD    & 0.388 & 0.444 & linear, raw field & unstable ($\max|\mu|$ 1.01--1.07) \\
FNO    & 0.019 & 0.157 & none & diverges ($t = 81$) \\
LRAN   & $0.048 \pm 0.001$ & $0.247 \pm 0.002$ & discrete $A$, unconstrained & unstable ($\max|\mu|$ 1.05 / 1.33--1.39) \\
KNO    & $0.016 \pm 0.001$ & $0.169 \pm 0.001$ & discrete $K$ in nonlinear maps & bounded, empirical \\
LGN-KM & 0.063 & 0.240 & continuous $L_k = S - D_k$ & $\mathrm{Re}(\lambda) \le 0$ guaranteed \\
\bottomrule
\end{tabular}
\end{table}

\subsection{Model transfer}

\begin{figure}[!t]
\begin{subfigure}{0.32\linewidth}
  \centering
  \includegraphics[height=4.5cm, keepaspectratio]{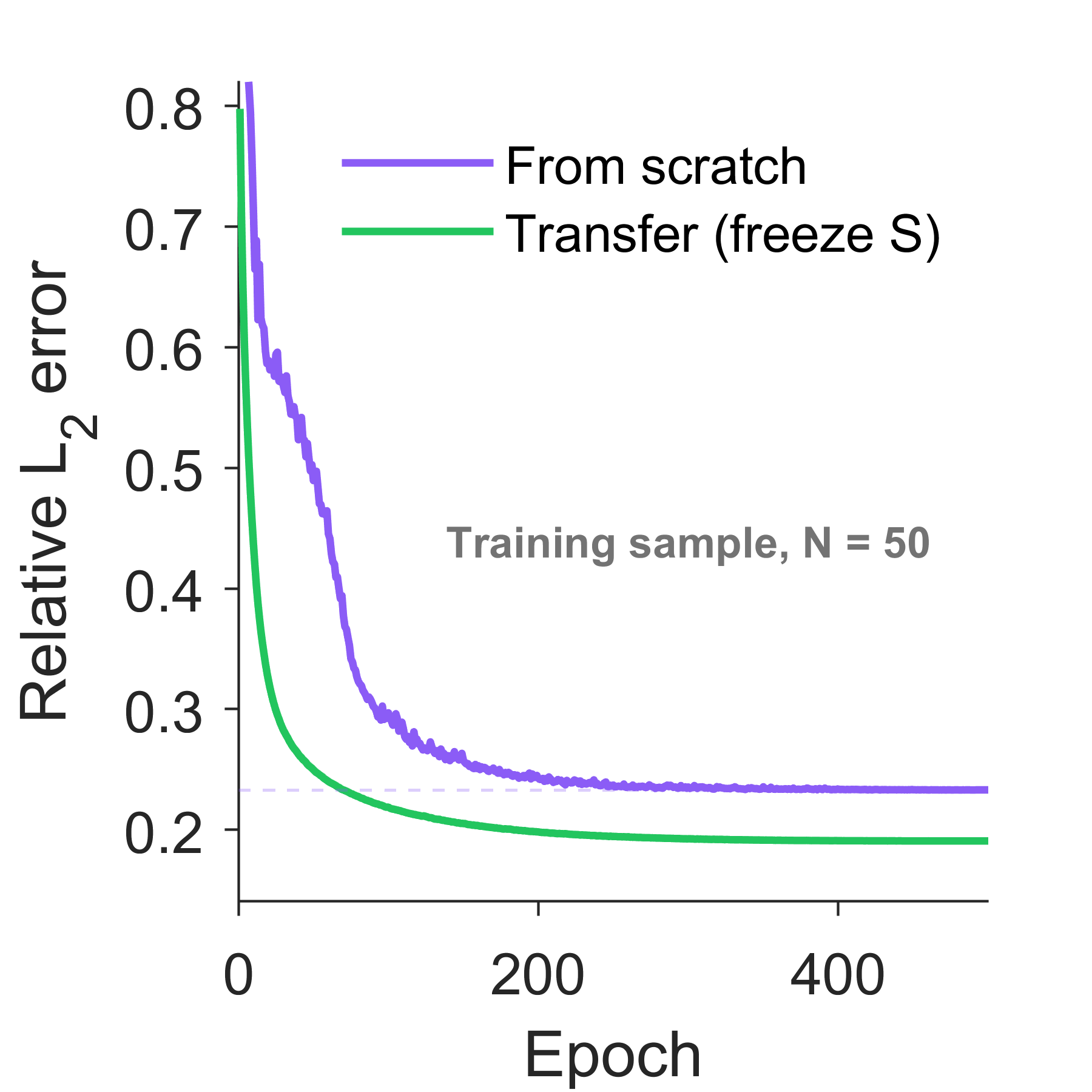}
  \caption{}
\end{subfigure}
\hfill
\begin{subfigure}{0.33\linewidth}
  \centering
  \includegraphics[height=4.5cm, keepaspectratio]{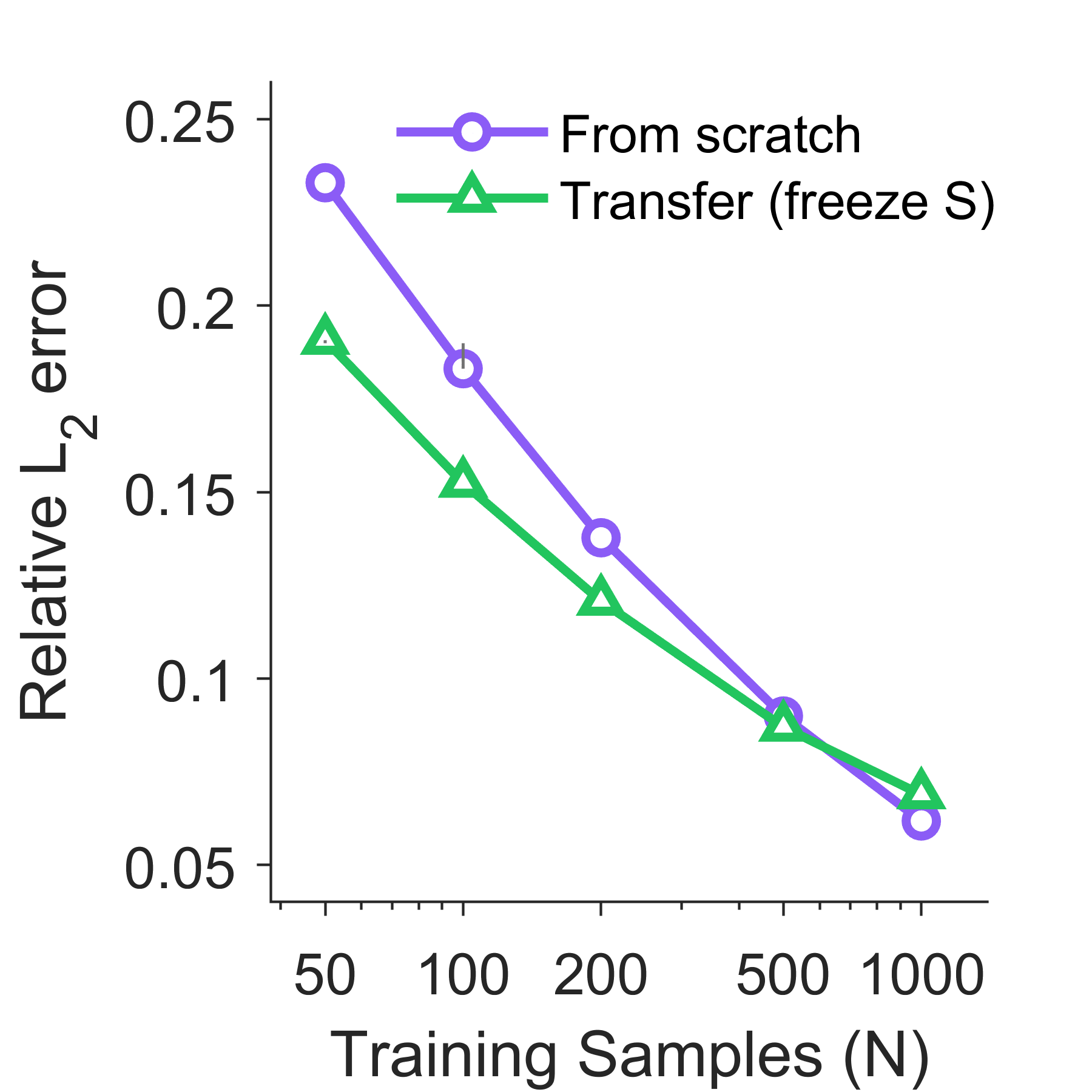}
  \caption{}
\end{subfigure}
\hfill
\begin{subfigure}{0.33\linewidth}
  \centering
  \includegraphics[height=4.5cm, keepaspectratio]{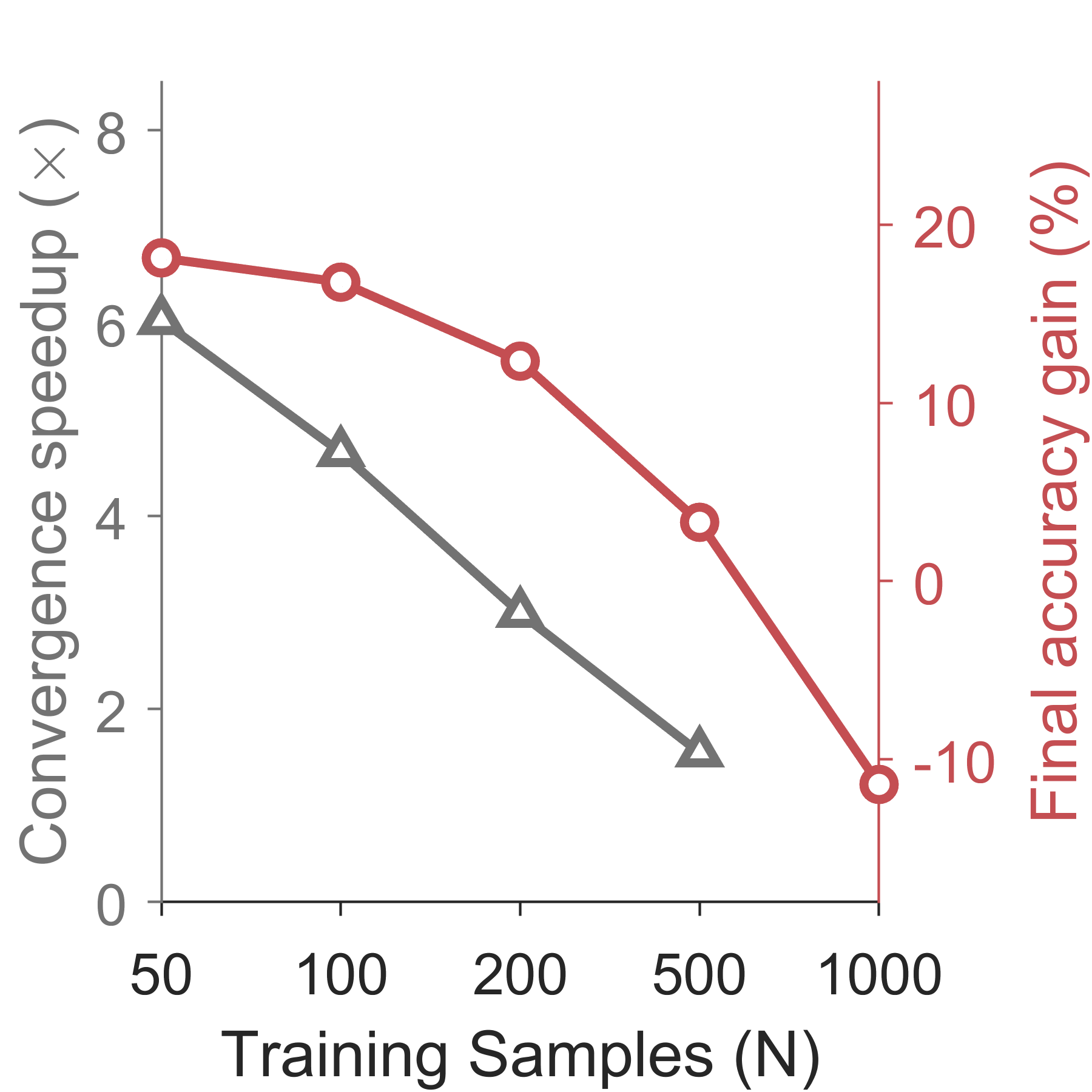}
  \caption{}
\end{subfigure}
\caption{Physics-informed cross-viscosity model transfer ($\nu = 10^{-5} \to \nu = 10^{-3}$). $S$ is frozen from the pretrained model; encoder, decoder, and $D$ are fine-tuned. (a) Learning curves at $N = 50$. (b) Test error across data budgets. (c) Convergence speedup (gray) and final accuracy gain (red) vs training sample size.}
\label{fig:transfer}
\end{figure}

Section~3.2 established that the singular value profile of $S$ is universal across viscosities ($R^2 = 0.870 \pm 0.063$) and that the Laplacian coefficient $\alpha$ adapts to the flow regime. This decomposition into universal and regime-specific components yields a direct prediction: the coupling matrix $S$ should transfer across viscosities without retraining, while only the dissipation parameters and the Koopman lifting require adaptation. We test this prediction by freezing $S$ from the $\nu = 10^{-5}$ model and fine-tuning the encoder, decoder, and dissipation parameters $(d, \alpha)$ on $\nu = 10^{-3}$ data at training sample size $N \in \{50, 100, 200, 500, 1000\}$. The frozen $S$ contributes the entire conservative coupling structure, carried from one flow regime to another without modification.

Figure~\ref{fig:transfer}(a) shows learning curves at $N = 50$, the most data-scarce condition. The transferred model reaches the scratch baseline's best test error at epoch $71$, a ${\sim}6\times$ convergence speedup, and converges to $18\%$ lower final error. Figure~\ref{fig:transfer}(b) shows the effect across all data budgets: transfer improves accuracy at every $N \le 500$, with the gain largest at low $N$ and reversing at $N = 1000$ ($7$--$11\%$ worse than scratch), where the target data alone is sufficient. Transfer at $N = 50$ approaches training from scratch at $N = 100$ ($0.191$ vs $0.183$). Figure~\ref{fig:transfer}(c) summarizes both the convergence speedup and the final accuracy gain as a function of $N$. Both decline monotonically from low to high data; the prior dominates when data is scarce and becomes unnecessary when data is abundant. This is physics-informed transfer rather than empirical transfer: the transferred object is $S$, whose normalized singular-value profile matched across the two regimes in Section~3.2. Fine-tuning $S$ yields ${<}5\%$ gain (Appendix Table~\ref{tab:transfer}), confirming $S$ does not require regime adaptation.

\section{Related work}

\paragraph{Neural operators.}
Fourier Neural Operators [1], DeepONet [2], and transformer-based variants [3] learn solution operators mapping initial conditions to future states. SGNO [8] constrains a diagonal generator to have non-positive real part for stable autoregressive rollouts. Its learned generator is structurally equivalent to the $D$-only ablation in Table~\ref{tab:ablation}; the architecture compensates with a separate nonlinear forcing network, but the generator itself carries no inter-channel coupling or oscillatory structure (Fig.~\ref{fig:ablation}c). The Koopman neural operator (KNO) [18] applies a discrete Koopman operator to the Fourier modes of a lifted field inside nonlinear observation maps that re-encode the field at every step, so the operator's spectrum does not by itself describe the predictor and carries no stability constraint.

\paragraph{Structure-preserving networks.}
Hamiltonian Neural Networks [9], Lagrangian Neural Networks [10], and SympNets [11] preserve energy or symplectic structure via numerical integration. Dissipative SymODEN [12] extends this to non-conservative systems through a port-Hamiltonian $(J - D)$ decomposition that separates conservative and dissipative components which is structurally analogous to our $S - D$. LGN-KM differs by acting on the generator rather than the energy function (direct eigenvalue access) and propagating via matrix exponentiation rather than numerical integration.

\paragraph{Deep Koopman methods.}
Lusch et al.\ [4], Yeung et al.\ [5], and Gin et al.\ [16], who learn global coordinate transformations that linearize PDEs, learn the discrete-time operator $K$ rather than the continuous-time generator $L$, and do not impose physical structure such as a stability constraint or a decomposition into conservative and dissipative components. Linearly recurrent autoencoders (LRAN) [17] learn a lift together with an unconstrained linear flow map $A$; its eigenvalues are per-step multipliers, and continuous-time rates require $\log(A)/\Delta t$, which is multivalued above the Nyquist limit. CT-KAE [7] decomposes a global Koopman generator into skew-symmetric and symmetric components. However, the symmetric part is unconstrained and stability is empirical rather than architectural. KoopGen [20] likewise splits a continuous-time generator into skew-adjoint and self-adjoint parts but carries no wavenumber structure. LGN-KM replaces the symmetric component with a positive-definite diagonal $D_k$ parameterized by wavenumber, guaranteeing $\mathrm{Re}(\lambda) \le 0$ by construction and providing per-mode spectral access that yields dispersion relations, dissipation scaling, and cross-regime transfer.

\paragraph{Dynamic mode decomposition.}
DMD [13, 19] extracts dynamic modes via a global linear operator on snapshot data. Extended DMD [14] enriches the observable space with nonlinear dictionary functions, but the dictionaries are manually selected and the resulting operator remains unconstrained. LGN-KM differs in three respects: (i) a learned nonlinear Koopman lifting trained end-to-end, (ii) stability guaranteed by $S{-}D$ where DMD/EDMD modes can be unstable, and (iii) per-Fourier-mode generators that yield wavenumber-dependent eigenvalues (dispersion relations) rather than a single global operator with no spatial frequency structure.

\section{Discussion and limitations}\label{sec:disc}

\paragraph{Latent representation and calibration.}
The recovered coefficients $d$ and $\alpha$ live in the learned latent basis. How closely the learned exponent tracks the viscous value, and the learned damping slopes track the diffusivities, measures how directly $d$ and $\alpha$ correspond to physical coefficients. This is the structure of standard physical measurement; it potentially enables measurement of dynamical parameters from trajectory data once calibrated against known cases.

\paragraph{Accuracy trade-off.}
LGN-KM is less accurate than FNO within the training horizon (relative $L_2 = 0.240 \pm 0.0003$ vs.\ $0.157$ [1] at $\nu = 10^{-5}$). The matched unconstrained generator reaches the same error (Appendix~\ref{app:ablation}), so the gap comes from the latent-linear architecture, a single lift propagated linearly and decoded pointwise, rather than from the $S{-}D$ constraint itself. For in-horizon prediction alone, a neural operator such as FNO is the better choice. LGN-KM is preferable when the generator itself is the object of interest, for reading decay rates and frequencies, checking a learned model against a known dispersion or damping law, transferring structure across regimes with little data, or querying arbitrary times stably.

\paragraph{Scalability.}
The matrix exponential costs $O(r^3)$ per Fourier mode, negligible at $r = 32$ but limiting for much larger latent dimensions. Krylov subspace [15] approximations could extend applicability to $r \sim 1000$. The per-mode independence of $L_k$ enables parallelization across all $M^2$ Fourier modes.

\paragraph{Designing the generator.}
Known structure enters in three places: the observable (on NLSE, the complex field exposes dispersion that the intensity alone does not), the basis (Fourier for periodic, cosine for no-flux), and the function class of $D_k$ from the PDE's dissipative operator. When that class is not known, the learnable-exponent experiment of Section~3.2 suggests a two-stage route: a low-parameter free form identifies the operator class, then a structured parameterization is fitted for the final readout.

\paragraph{Scope.}
We demonstrate on 2D periodic Navier--Stokes, 2D FitzHugh--Nagumo with periodic and no-flux boundaries (Appendix~\ref{app:nonperiodic}), and 1D fiber propagation. The $S{-}D$ architecture applies to any basis that diagonalizes the dissipative operator under the domain's boundary conditions; Chebyshev bases for general Dirichlet conditions, 3D domains, and multi-physics systems are not tested here. Furthermore, since the symmetric part of $L_k$ is $-D_k \preceq 0$, the latent norm can only decrease. Therefore, the latent dynamics cannot sustain energy injection, so long-horizon predictions of forced flows decay to the mean (Section 4.1), and any transient growth of the physical field is carried by the nonlinear encoder and decoder rather than by the generator. An input term $\dot{z} = Lz + Bu$ is the natural extension and is not evaluated here.

\paragraph{Gauge freedom.}
The Koopman lifting is not unique as different encoders produce different latent coordinates with different $S$ matrices, while the gauge-invariant spectral structure is preserved. Resolving this freedom via shared encoders across regimes, canonical forms, or gauge-alignment losses is a natural direction for enabling direct parameter transfer without fine-tuning.

\section{Conclusion}

We introduced LGN-KM, a neural operator that learns the structured Koopman generator of nonlinear PDE dynamics through the decomposition $L_k = S - D_k$. On 2D Navier--Stokes turbulence, this opens linear-systems analysis to a nonlinear PDE: the generator yields a complete multi-branch dispersion relation, a learned dissipation exponent of 2.2--2.4, and matched spectral structure across models trained independently at viscosities $100\times$ apart. On pulse propagation in optical fiber, the oscillation frequencies of the generator match the fiber's dispersion relation with no fitted constant. The same architectural constraint that enables these diagnostics guarantees long-horizon stability, provides $O(1)$ continuous-time evaluation, and supports physics-informed cross-regime transfer. The generator trades short-horizon accuracy for direct spectral access to the learned dynamics, complementing neural operators optimized for prediction.

\begin{ack}
This work was partially supported by the Department of Energy Grant No.\ DE-SC0022248, Office of Naval Research Grant No.\ N00014-21-1-2634, and Air Force Office of Scientific Research Grants No.\ FA9550-21-1-0305 and FA9550-22-1-0433. The authors declare no competing interests.
\end{ack}

\section*{References}

\medskip
{\small

[1] Z.\ Li et al., ``Fourier Neural Operator for Parametric Partial Differential Equations,'' ICLR 2021, arXiv:2010.08895, Oct. 2020.

[2] L.\ Lu, P.\ Jin, G.\ Pang, Z.\ Zhang, and G.\ E.\ Karniadakis, ``Learning nonlinear operators via DeepONet based on the universal approximation theorem of operators,'' \textit{Nat.\ Mach.\ Intell.}, vol.\ 3, no.\ 3, pp.\ 218--229, Mar.\ 2021, doi: 10.1038/s42256-021-00302-5.

[3] Z.\ Hao et al., ``GNOT: A General Neural Operator Transformer for Operator Learning,'' Jun.\ 2023.

[4] B.\ Lusch, J.\ N.\ Kutz, and S.\ L.\ Brunton, ``Deep learning for universal linear embeddings of nonlinear dynamics,'' \textit{Nat.\ Commun.}, vol.\ 9, no.\ 1, p.\ 4950, Nov.\ 2018, doi: 10.1038/s41467-018-07210-0.

[5] E.\ Yeung, S.\ Kundu, and N.\ Hodas, ``Learning Deep Neural Network Representations for Koopman Operators of Nonlinear Dynamical Systems,'' Nov.\ 2017.

[6] S.\ Jamil and R.\ Kapadia, ``Interpretable Physics Extraction from Data for Linear Dynamical Systems using Lie Generator Networks,'' Mar.\ 2026, arXiv:2603.27442.

[7] R.\ Grozavescu, P.\ Zhang, M.\ Girolami, and E.\ Meunier, ``Towards Efficient and Stable Ocean State Forecasting: A Continuous-Time Koopman Approach,'' Mar.\ 2026, arXiv:2603.05560.

[8] J.\ Li, P.\ Jiang, H.\ Saleem, Z.\ Wang, P.\ Koniusz, and F.\ D.\ Salim, ``SGNO: Spectral Generator Neural Operators for Stable Long Horizon PDE Rollouts,'' arXiv:2602.18801v2, May 2026.

[9] S.\ Greydanus, M.\ Dzamba, and J.\ Yosinski, ``Hamiltonian Neural Networks,'' Sep.\ 2019.

[10] M.\ Cranmer, S.\ Greydanus, S.\ Hoyer, P.\ Battaglia, D.\ Spergel, and S.\ Ho, ``Lagrangian Neural Networks,'' Jul.\ 2020.

[11] P.\ Jin, Z.\ Zhang, A.\ Zhu, Y.\ Tang, and G.\ E.\ Karniadakis, ``SympNets: Intrinsic structure-preserving symplectic networks for identifying Hamiltonian systems,'' Aug.\ 2020.

[12] Y.\ D.\ Zhong, B.\ Dey, and A.\ Chakraborty, ``Dissipative SymODEN: Encoding Hamiltonian Dynamics with Dissipation and Control into Deep Learning,'' ICLR 2020 Workshop on Integration of Deep Neural Models and Differential Equations, arXiv:2002.08860, Feb. 2020.

[13] P.\ J.\ Schmid, ``Dynamic mode decomposition of numerical and experimental data,'' \textit{J.\ Fluid Mech.}, vol.\ 656, pp.\ 5--28, Aug.\ 2010, doi: 10.1017/S0022112010001217.

[14] M.\ O.\ Williams, I.\ G.\ Kevrekidis, and C.\ W.\ Rowley, ``A Data-Driven Approximation of the Koopman Operator: Extending Dynamic Mode Decomposition,'' Aug.\ 2014, doi: 10.1007/s00332-015-9258-5.

[15] Y.\ Saad, ``Analysis of Some Krylov Subspace Approximations to the Matrix Exponential Operator,'' \textit{SIAM J.\ Numer.\ Anal.}, vol.\ 29, no.\ 1, pp.\ 209--228, Feb.\ 1992, doi: 10.1137/0729014.

[16] C.\ Gin, B.\ Lusch, S.\ L.\ Brunton, and J.\ N.\ Kutz, ``Deep learning models for global coordinate transformations that linearise PDEs,'' \textit{Eur.\ J.\ Appl.\ Math.}, vol.\ 32, no.\ 3, pp.\ 515--539, 2021.

[17] S.\ E.\ Otto and C.\ W.\ Rowley, ``Linearly recurrent autoencoder networks for learning dynamics,'' \textit{SIAM J.\ Appl.\ Dyn.\ Syst.}, vol.\ 18, no.\ 1, pp.\ 558--593, 2019.

[18] W.\ Xiong, X.\ Huang, Z.\ Zhang, R.\ Deng, P.\ Sun, and Y.\ Tian, ``Koopman neural operator as a mesh-free solver of non-linear partial differential equations,'' \textit{J.\ Comput.\ Phys.}, vol.\ 513, 113194, 2024.

[19] J.\ H.\ Tu, C.\ W.\ Rowley, D.\ M.\ Luchtenburg, S.\ L.\ Brunton, and J.\ N.\ Kutz, ``On dynamic mode decomposition: Theory and applications,'' \textit{J.\ Comput.\ Dyn.}, vol.\ 1, no.\ 2, pp.\ 391--421, 2014.

[20] L.\ Su, J.\ Shu, R.\ Liu, D.\ Meng, and Z.\ Xu, ``KoopGen: Koopman generator networks for representing and predicting dynamical systems with continuous spectra,'' arXiv:2602.14011, Feb.\ 2026.
}

\newpage
\appendix
\renewcommand{\thesection}{A.\arabic{section}}
\renewcommand{\thetable}{A.\arabic{table}}
\renewcommand{\thefigure}{A.\arabic{figure}}
\setcounter{table}{0}
\setcounter{figure}{0}

\section{Training details}\label{app:training}

\paragraph{Dataset.}
We use the 2D Navier--Stokes benchmark from Li et al.\ [1]: vorticity fields on a $64 \times 64$ periodic domain. At $\nu = 10^{-5}$ ($\mathrm{Re} \approx 10^5$), the dataset contains $1200$ trajectories of length $T = 20$; we use $1000$ for training and $200$ for testing with $T_\mathrm{in} = 10$ input frames and $T_\mathrm{out} = 10$ predicted frames. At $\nu = 10^{-3}$ ($\mathrm{Re} \approx 10^3$), the dataset contains $5000$ trajectories of length $T = 50$; we use $1000/200$ with $T_\mathrm{in} = 10$ and $T_\mathrm{out} = 40$. No normalization is applied to the data.

\paragraph{LGN-KM.}
Architecture: $r = 32$ latent channels, $M = 12$ retained Fourier modes per dimension. The encoder uses two FNO-style spectral convolution blocks (spectral convolution $+$ pointwise residual $+$ GELU) to lift from $T_\mathrm{in} + 2$ input channels to $r = 32$ latent channels. The decoder is pointwise ($r \to 128 \to 1$) with no spatial mixing, so all spatial dynamics are forced through the generator. The generator has $560$ interpretable parameters: $496$ in $S$ and $64$ in $D$ (32 base damping $d$, 32 Laplacian coefficients $\alpha$). Optimizer: AdamW (lr $= 10^{-3}$, weight decay $10^{-4}$) with cosine annealing over $500$ epochs. Batch size $10$. Gradient clipping at norm $1.0$. All models train in under two hours on a single NVIDIA RTX A5000 GPU.

\paragraph{FNO-2D baseline.}
Four spectral convolution layers, width $20$, $M = 12$ modes, followed by a two-layer pointwise head ($20 \to 128 \to 1$). Optimizer: Adam (lr $= 10^{-3}$, weight decay $10^{-4}$) with step decay ($\times 0.5$ every $100$ epochs) over $500$ epochs. Batch size $20$. Autoregressive rollout during both training and evaluation. Our reimplementation reaches relative $L_2 = 0.157$ at $\nu = 10^{-5}$, close to the reported value $0.1556$ [1].

\paragraph{Metric.}
Per-sample relative $L_2$ error over the full predicted volume ($T_\mathrm{out} \times H \times W$), averaged over test samples:
\begin{equation*}
\frac{1}{N} \sum_i \frac{\|\hat{u}_i - u_i\|}{\|u_i\|}.
\end{equation*}

\section{Effect of channel dimension \texorpdfstring{$r$}{r}}\label{app:channel}

\begin{figure}[!h]
\centering
\includegraphics[width=0.4\linewidth]{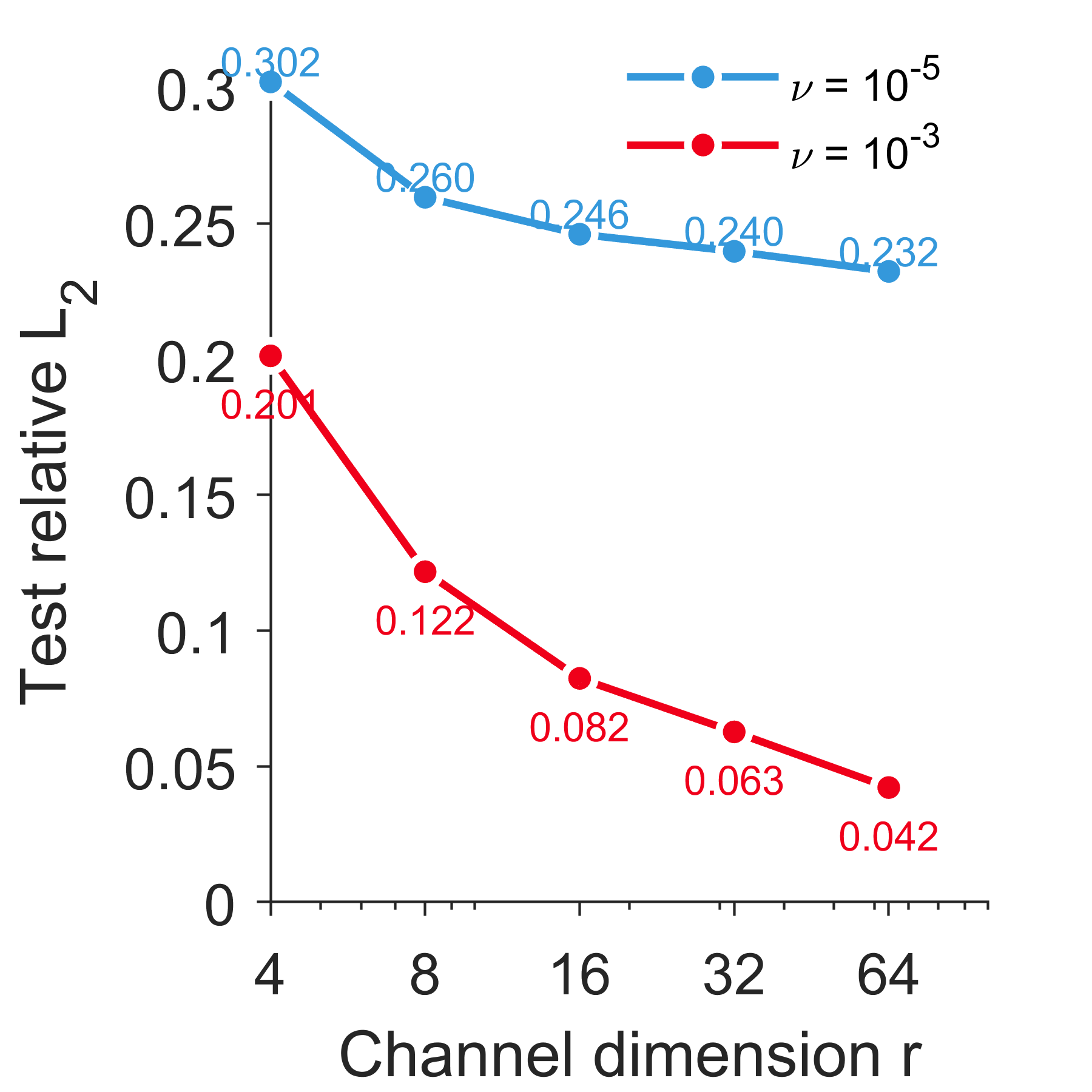}
\caption{Test error vs channel dimension ($r$).}
\label{fig:channel}
\end{figure}

We train LGN-KM at $r \in \{4, 8, 16, 32, 64\}$ channels at both viscosities with all other hyperparameters fixed ($M = 12$, $w = 32$, $500$ epochs). Figure~\ref{fig:channel} illustrates that prediction accuracy improves monotonically with $r$ at both viscosities. At $\nu = 10^{-3}$, test $L_2$ drops from $0.201$ ($r = 4$) to $0.042$ ($r = 64$). At $\nu = 10^{-5}$, the reduction is smaller: $0.302$ to $0.232$. The accuracy gap between viscosities persists at all $r$, consistent with the higher intrinsic complexity of $\mathrm{Re} \approx 10^5$ turbulence. At every $r$, the dominant eigenvalue of the generator (the slowest-decaying mode at each wavevector) maintains the viscosity dissipation scaling with wavenumber ($\mathrm{Re}(\lambda)$ vs $|k|^2$, $R^2 > 0.88$ for $r \le 32$) at both viscosities. At $r = 64$ the dominant-branch $R^2$ drops to $0.756$ for $\nu = 10^{-5}$, though the median branch $R^2$ remains $0.990$, indicating redistribution across channels rather than loss of physics. The main text uses $r = 32$.

\section{Model transfer}\label{app:transfer}

Table~\ref{tab:transfer} compares two training strategies on $\nu = 10^{-3}$ data using a $\nu = 10^{-5}$ pretrained model: training while freezing $S$ and fine-tuning encoder/decoder/$D$ (the method reported in the main text), and fine-tuning all parameters including $S$.

\begin{table}[!h]
\caption{Cross-viscosity transfer ($\nu = 10^{-5} \to 10^{-3}$) test $L_2$ error. $\Delta$ = gap between freeze-$S$ and transfer-all as $\%$ of transfer-all error.}
\label{tab:transfer}
\centering
\begin{tabular}{rccc}
\toprule
Train Sample, $N$ & Freeze $S$ & Transfer All & $\Delta$ ($\%$) \\
\midrule
50 & 0.191 & 0.182 & 4.5 \\
100 & 0.152 & 0.147 & 4.0 \\
200 & 0.121 & 0.118 & 2.5 \\
500 & 0.087 & 0.085 & 2.0 \\
1000 & 0.069 & 0.066 & 4.0 \\
\bottomrule
\end{tabular}
\end{table}

Freeze-$S$ and transfer-all perform within $5\%$ at every $N$, confirming that $S$ does not require adaptation to the target regime. At $N = 1000$ both transfer strategies are $7$--$11\%$ worse than training from scratch ($0.062$), consistent with the prior becoming unnecessary when target data is sufficient.

\section{Cross-domain application}\label{app:fhn}

To evaluate the generality of the $S{-}D$ decomposition beyond fluid mechanics, we apply LGN-KM with no architecture modification to the 2D FitzHugh--Nagumo equations, a canonical model of excitable media used across neuroscience, cardiac dynamics, and chemical pattern formation:
\begin{align*}
\partial u / \partial t &= D_u \nabla^2 u + u - u^3/3 - v, \\
\partial v / \partial t &= D_v \nabla^2 v + \varepsilon (u + a - bv),
\end{align*}
where $u$ is the activator (observed), $v$ is the inhibitor (hidden), and $D_u, D_v$ are diffusion coefficients. The model observes only $u$ while the inhibitor dynamics are absorbed into the Koopman lifting via the $T_\mathrm{in} = 10$ input frames. The nonlinearity here is cubic ($u^3/3$), structurally distinct from the quadratic advection $(u \cdot \nabla)\omega$ of Navier--Stokes.

\begin{figure}[!t]
\begin{subfigure}{0.32\linewidth}
  \centering
  \includegraphics[height=4.5cm, keepaspectratio]{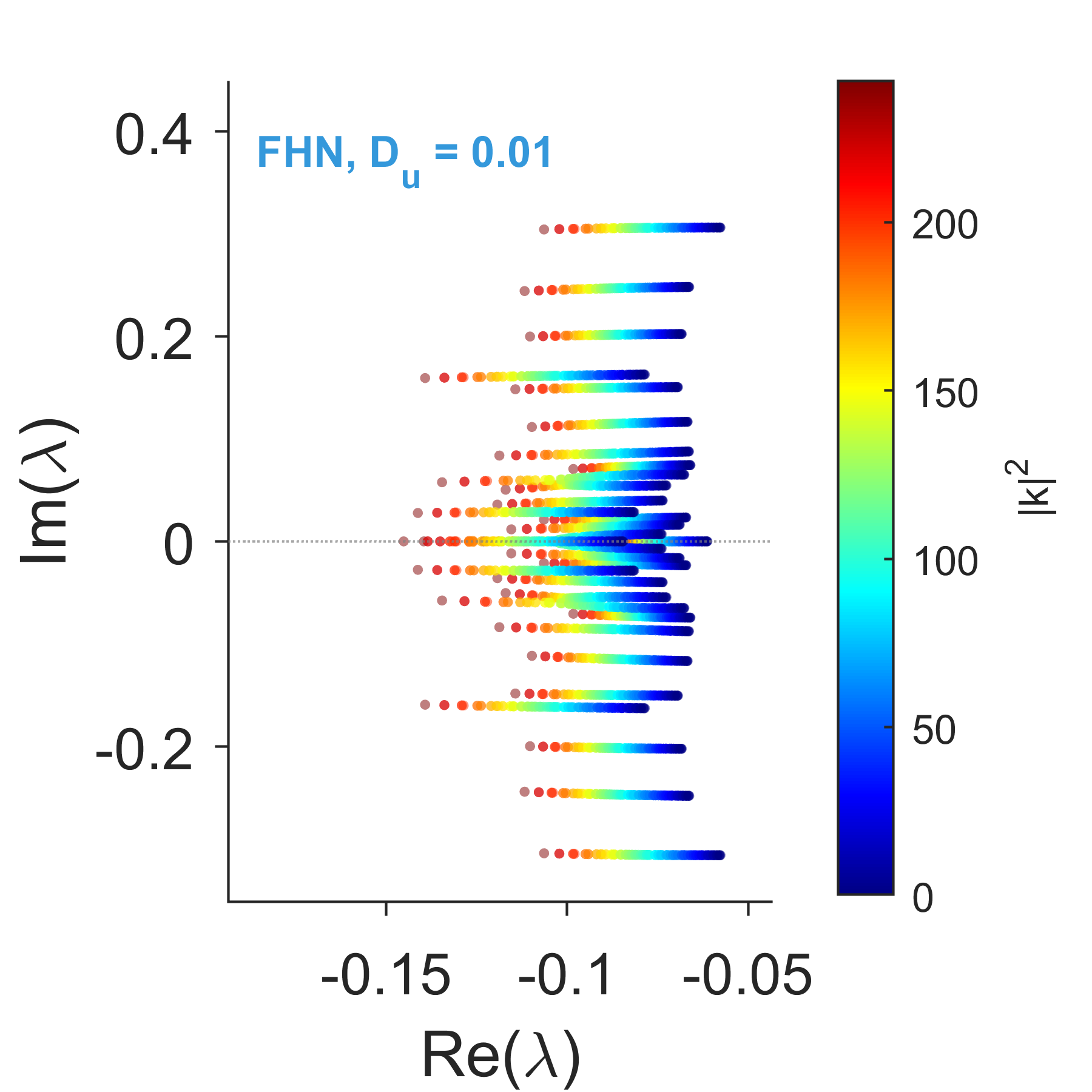}
  \caption{}
\end{subfigure}
\hfill
\begin{subfigure}{0.33\linewidth}
  \centering
  \includegraphics[height=4.5cm, keepaspectratio]{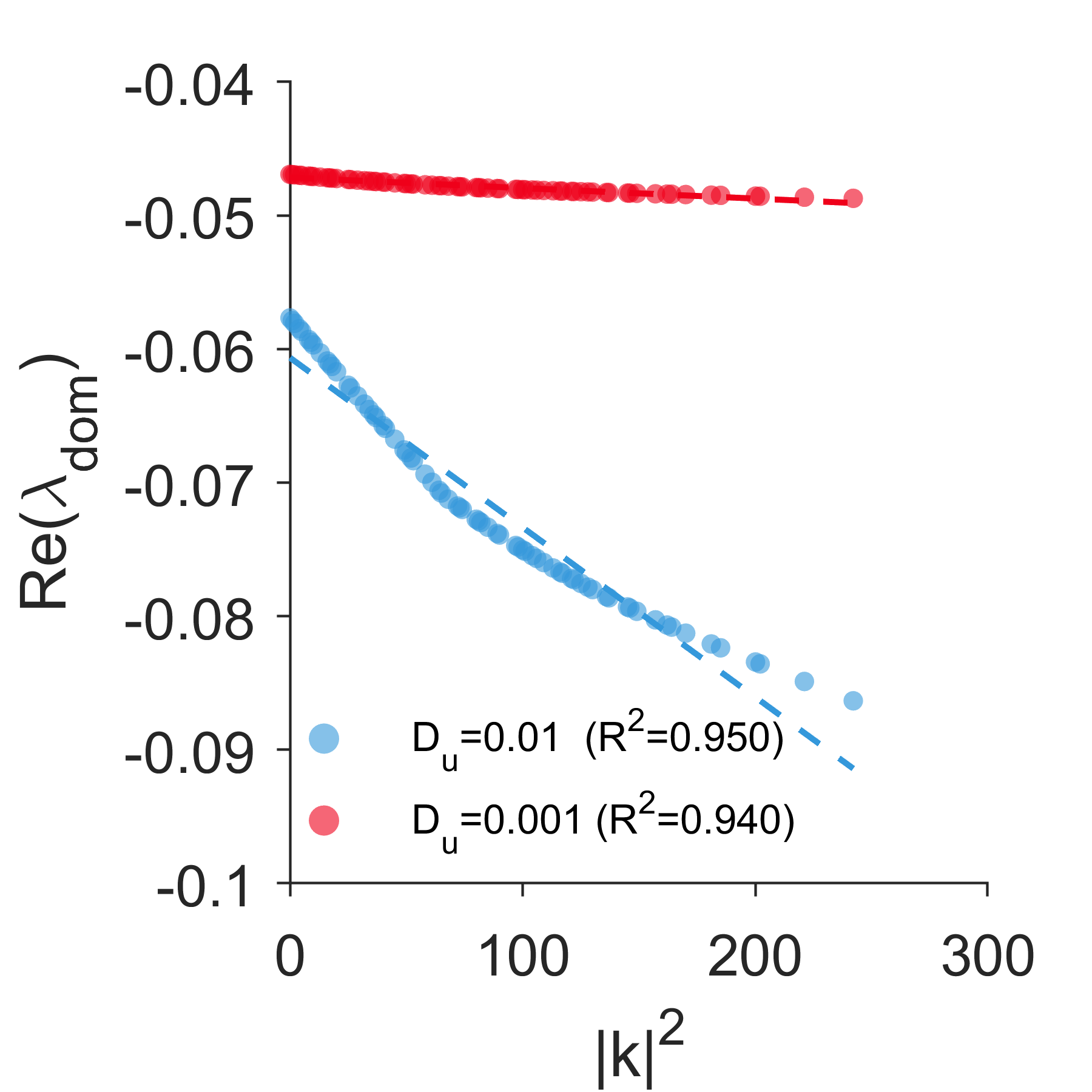}
  \caption{}
\end{subfigure}
\hfill
\begin{subfigure}{0.33\linewidth}
  \centering
  \includegraphics[height=4.5cm, keepaspectratio]{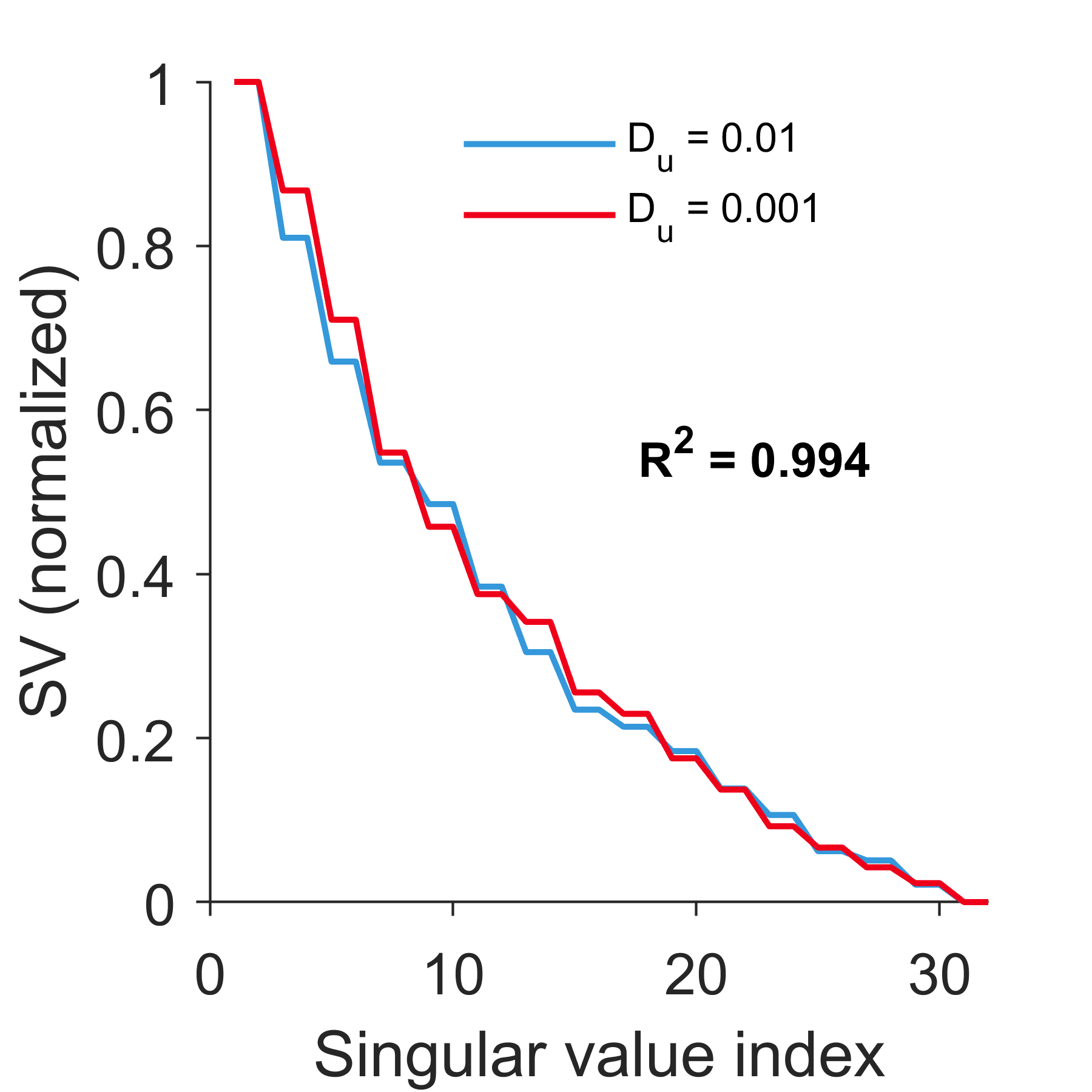}
  \caption{}
\end{subfigure}
\caption{Generalization to 2D FitzHugh--Nagumo reaction-diffusion with no architecture modification. (a) Eigenspectra at diffusion coefficient $D_u = 0.01$. (b) $\mathrm{Re}(\lambda_{\mathrm{dom}})$ vs $|k|^{2}$ recovers diffusive scaling at both diffusion coefficients ($R^{2} > 0.93$); higher $D_u$ produces steeper decay. (c) Normalized SVD profiles of $S$ match across diffusion coefficients ($R^{2} = 0.994$), replicating the universal coupling structure found in Navier--Stokes (Fig.~\ref{fig:universality}b).}
\label{fig:fhn}
\end{figure}

The $S{-}D$ architecture predicts a specific decomposition for this system: reaction kinetics ($u - u^3/3 - v$) contain no spatial derivatives and should be absorbed into the $k$-independent part of the generator, while diffusion ($D_u \nabla^2 u$) acts as $-D_u |k|^2$ in Fourier space and should appear in $D_k$. We train at two diffusion coefficients ($D_u = 0.01$ and $D_u = 0.001$, $10\times$ apart) on $64 \times 64$ periodic domains with $1200$ trajectories each, using identical architecture and hyperparameters as the Navier--Stokes experiments ($r = 32$, $M = 12$, $500$ epochs). Parameters: $D_v = 0.005$, $\varepsilon = 0.01$, $a = 0.7$, $b = 0.8$, domain $[0, 2\pi]^2$, $1200$ trajectories of length $T = 30$ at each $D_u$.

The generator recovers the expected structure. Figure~\ref{fig:fhn}(a) shows the eigenspectra at $D_u = 0.01$. Eigenvalues organize into vertical columns with $|k|^2$-dependent horizontal spread. The vertical column structure, in contrast to the Navier--Stokes spectra (Fig.~\ref{fig:eigenspectra}), reflects the fact that reaction kinetics contain no spatial derivatives, producing mode-independent imaginary parts. All $4{,}608$ eigenvalues satisfy $\mathrm{Re}(\lambda) < 0$. Figure~\ref{fig:fhn}(b) confirms that the dominant eigenvalue scales as $\mathrm{Re}(\lambda) \propto |k|^2$ at both diffusion coefficients ($R^2 = 0.950$ and $0.940$), with higher $D_u$ producing a steeper decay.

The universality analysis from Section~3.2 replicates on this system. Models trained independently at $D_u = 0.01$ and $D_u = 0.001$ learn in different latent coordinates ($S$ cosine similarity $0.449$; single seed, shared initialization), yet their $S$ singular value profiles match at $R^2 = 0.994$ (Fig.~\ref{fig:fhn}(c)), sorted base damping distributions match at $R^2 = 0.945$, and Laplacian coefficients $\alpha$ show the greatest divergence ($R^2 = 0.820$). The architecture recovers a physically consistent structure on a qualitatively different PDE, with a different nonlinearity, from a different physical domain, without modification.

\section{Structural ablation}\label{app:ablation}

To isolate the contribution of each component in the $S{-}D$ decomposition, we train three ablated variants on the $\nu = 10^{-5}$ dataset with identical encoder, decoder, optimizer, and hyperparameters.

\paragraph{Unconstrained $L$.}
The generator is $L_k = A + \mathrm{diag}(b \cdot |k|^2 / k_\mathrm{max}^2)$, where $A \in \mathbb{R}^{r \times r}$ is an unconstrained matrix (no skew-symmetry) and $b \in \mathbb{R}^r$ is unconstrained (no positivity enforcement). This preserves the per-mode Koopman structure and $k$-dependent diagonal but removes all structural constraints.

\paragraph{$S$-only ($D = 0$).}
The generator is $L_k = S$ for all $k$, with no dissipation parameters; all eigenvalues are purely imaginary.

\paragraph{$D$-only ($S = 0$).}
The generator is $L_k = -D_k$, a positive diagonal with no inter-channel coupling; all eigenvalues are real and negative.

\begin{figure}[!t]
\begin{subfigure}{0.32\linewidth}
  \centering
  \includegraphics[height=4.5cm, keepaspectratio]{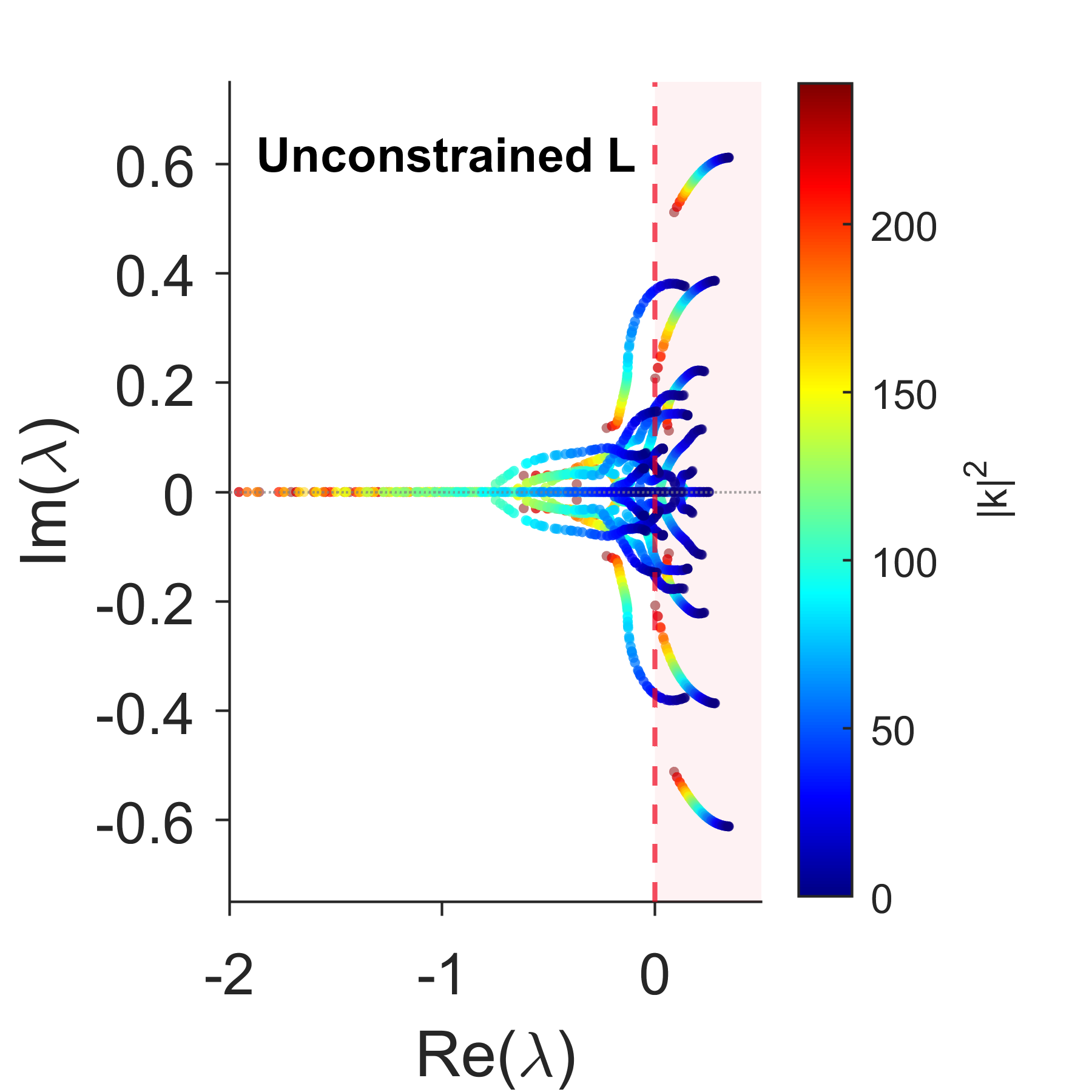}
  \caption{}
\end{subfigure}
\hfill
\begin{subfigure}{0.33\linewidth}
  \centering
  \includegraphics[height=4.5cm, keepaspectratio]{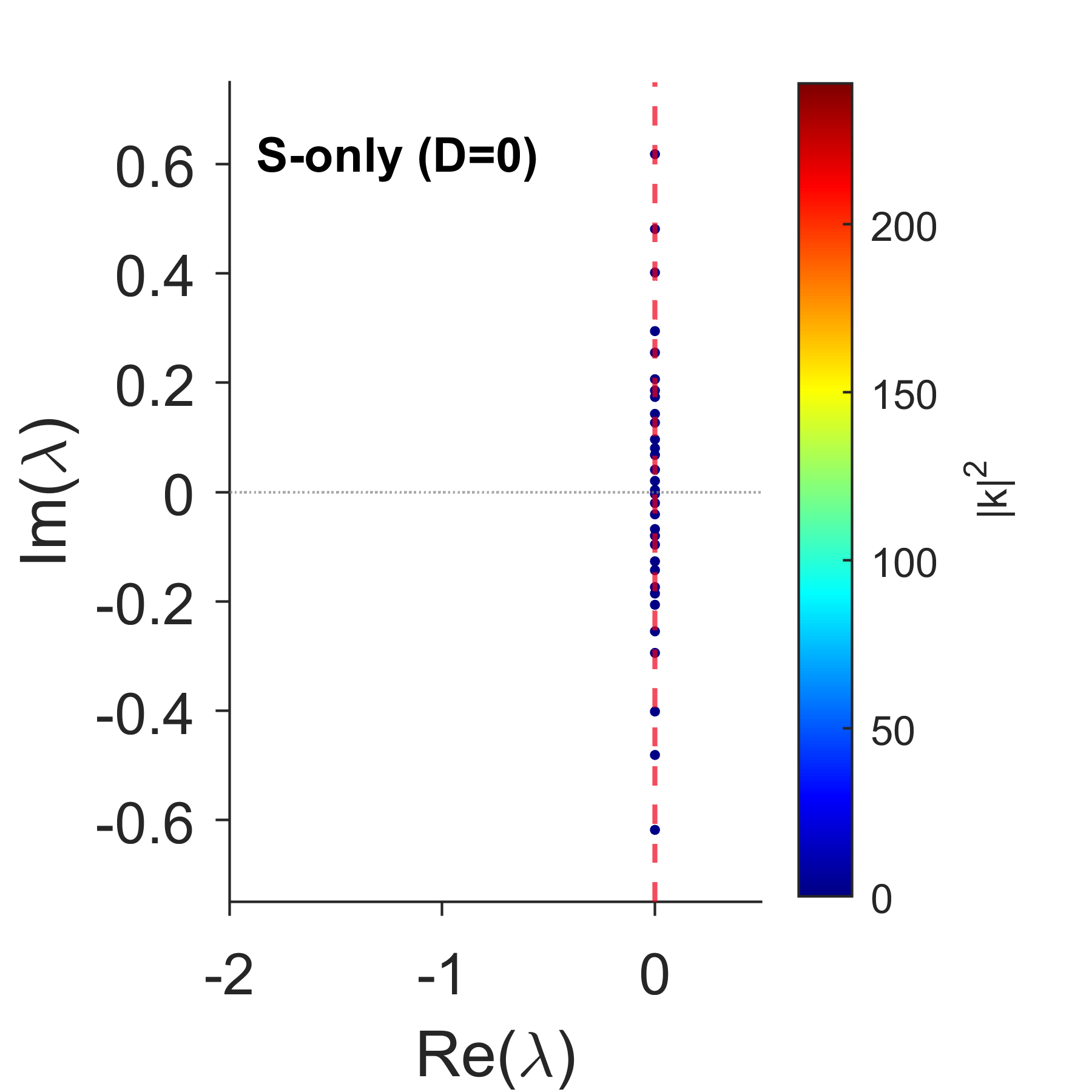}
  \caption{}
\end{subfigure}
\hfill
\begin{subfigure}{0.33\linewidth}
  \centering
  \includegraphics[height=4.5cm, keepaspectratio]{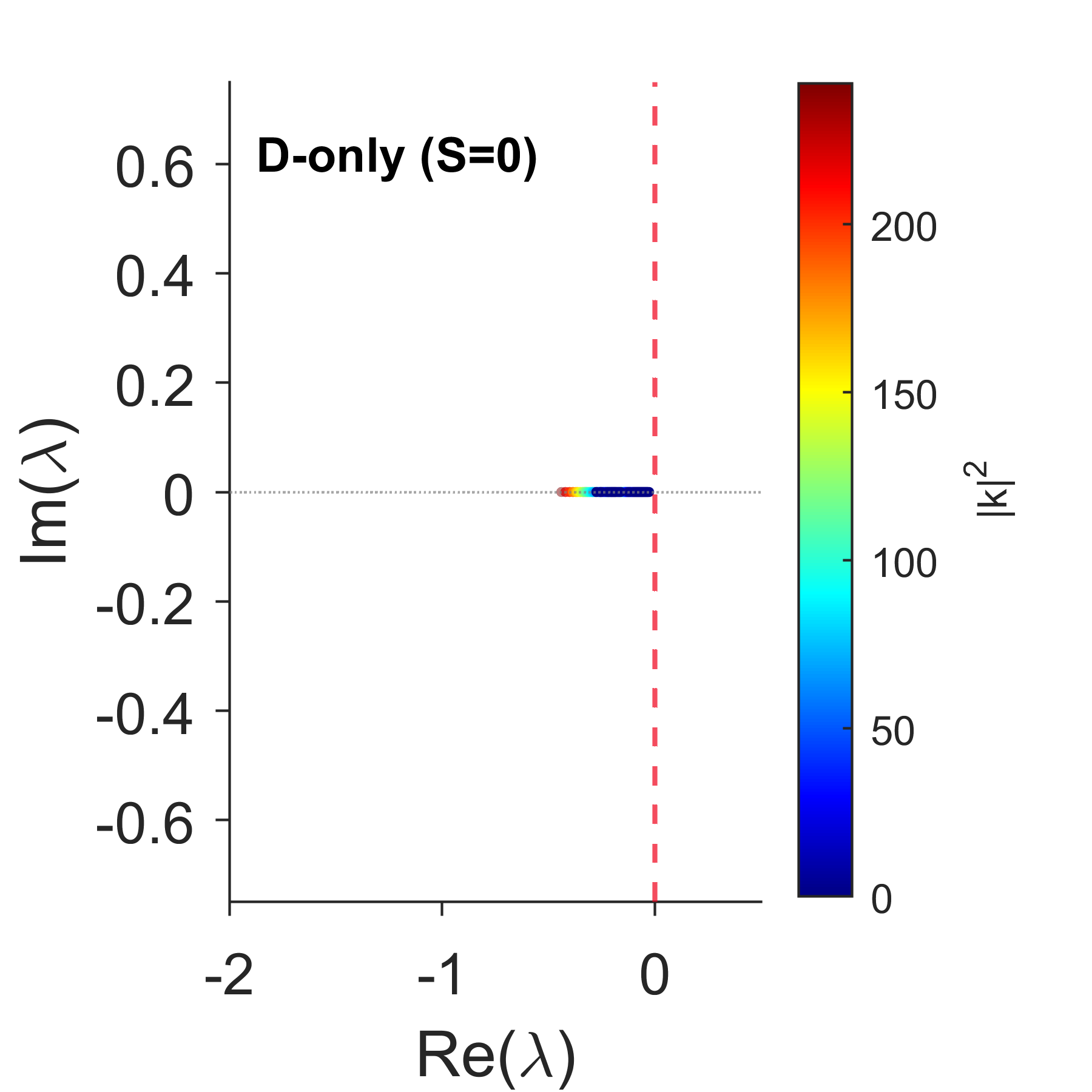}
  \caption{}
\end{subfigure}
\caption{Ablated eigenspectra ($\nu = 10^{-5}$). (a) Unconstrained $L$ produces dispersion branches but $41\%$ of eigenvalues are unstable ($\mathrm{Re} > 0$, shaded). (b) $S$-only collapses to the imaginary axis. (c) $D$-only to the real axis. Only $S{-}D$ yields stable, oscillatory, $k$-dependent structure simultaneously.}
\label{fig:ablation}
\end{figure}

\begin{table}[!h]
\caption{Structural ablation on $\nu = 10^{-5}$. All models share the same encoder, decoder, and training configuration ($r = 32$, $M = 12$, $500$ epochs).}
\label{tab:ablation}
\centering
\begin{tabular}{lcccc}
\toprule
Feature & $S{-}D$ & Unconstrained $L$ & $S$-only ($D{=}0$) & $D$-only ($S{=}0$) \\
\midrule
Test $L_2$ error & $0.240 \pm 0.0003$ & $0.240$ & $0.240$ & $0.253$ \\
Stability ($\mathrm{Re}(\lambda) < 0$) & Stable & Unstable & Marginal & Stable \\
Max $\mathrm{Re}(\lambda)$ & $-0.021$ & $+0.347$ & $0$ & $-0.026$ \\
$\mathrm{Im}(\lambda)$ range & $\pm 0.70$ & $\pm 0.61$ & $\pm 0.62$ & $0$ \\
Dissipation $R^2$ & $0.92$ & $0.94$ & N/A & $0.80$ \\
\bottomrule
\end{tabular}
\end{table}

The unconstrained generator is unstable at every training epoch, with $\max \mathrm{Re}(\lambda)$ reaching $+0.35$. This confirms that the $S{-}D$ constraint imposes no accuracy penalty while guaranteeing stability by construction. The $S$-only model matches accuracy, demonstrating that the encoder and decoder can absorb short-horizon dissipation. However, it produces a single generator for all modes (no $k$-dependence), eliminating the dispersion relation structure. The $D$-only model has the worst accuracy ($0.253$) and, as a diagonal generator, produces only real eigenvalues with no oscillatory dynamics and no inter-channel coupling structure to transfer across regimes. Only the full $S{-}D$ decomposition simultaneously provides matching accuracy, guaranteed stability, physically interpretable eigenspectra, and transferable coupling structure.

\section{Learnable exponent and free diagonal}\label{app:exponent}

Two variations of $D_k$ are trained under the paper protocol ($r = 32$, $M = 12$, $500$ epochs, three seeds per viscosity), keeping $S$ skew and $D_k$ positive so Proposition~\ref{prop:stab} holds in both. First, we parameterize $D_k = \mathrm{diag}\!\left(\mathrm{softplus}(d) + \mathrm{softplus}(\alpha)\cdot |k|^{p}/k_{\max}^{p}\right)$, with $p = \mathrm{softplus}(p_\mathrm{raw})$, initialized so $p = 1$.

\begin{table}[!h]
\caption{Learnable damping exponent.}
\label{tab:exponent}
\centering
\begin{tabular}{lccc}
\toprule
$\nu$ & $p$ (seeds 0/1/2) & test $L_2$ & fixed $p = 2$ \\
\midrule
$10^{-5}$ & 2.377 / 2.388 / 2.419 & 0.2407 / 0.2408 / 0.2406 & 0.2397 \\
$10^{-3}$ & 2.281 / 2.033 / 2.166 & 0.0615 / 0.0617 / 0.0617 & 0.0627 \\
\bottomrule
\end{tabular}
\end{table}

The exponent depends on the prediction horizon: at $\nu = 10^{-3}$, two earlier runs with $T_\mathrm{out} = 10$ instead of $40$ ended at $p \approx 1.2$, while the $T_\mathrm{out} = 40$ runs reported above end near $2.2$. We therefore read the result as support for near-quadratic damping within the tested family at the paper's horizons, not as an estimate of $p$ with a stated uncertainty.

Next, we parameterize $D_k = \mathrm{diag}(\mathrm{softplus}(B_k))$ with $B \in \mathbb{R}^{M \times M \times r}$ free, no functional form in $k$. Test $L_2$ is $0.2369$ / $0.2385$ / $0.2397$ at $\nu = 10^{-5}$ and $0.0598$ / $0.0523$ / $0.0589$ at $\nu = 10^{-3}$, matching or slightly below the fixed form. Averaged over wavenumbers, the learned dissipation increases with $|k|$ (Spearman $\rho \approx 0.95$ at $\nu = 10^{-5}$; $0.70$--$0.73$ per wavevector at $\nu = 10^{-3}$). An exponent fitted to this free diagonal is not stable. At $\nu = 10^{-5}$ an offset power-law fit gives $p \approx 3.3$--$5.2$ over $|k|^2 \le 20$, $1.8$--$2.6$ over $|k|^2 \le 60$, and $1.0$--$1.4$ over all modes, and $1.0$--$1.6$ at $\nu = 10^{-3}$. The free diagonal therefore reproduces the accuracy and the monotone trend of the imposed form but does not identify the exponent, which is why the main model keeps the parametric form.

\section{Non-periodic domain}\label{app:nonperiodic}

The Fourier lift assumes spatial periodicity. To test the architecture on a non-periodic domain, we repeat the FitzHugh--Nagumo experiment of Appendix~\ref{app:fhn} with homogeneous Neumann (no-flux) boundary conditions, the physically natural choice for reaction--diffusion in a bounded domain, and replace the FFT by a type-II discrete cosine transform. Every cosine basis function has zero wall-normal derivative, so the no-flux condition is satisfied exactly, and the pointwise polynomial kinetics of the model preserve the cosine subspace. $D_k$ is indexed by the cosine wavenumbers $|k_{m,n}|^2 = (\pi m/L)^2 + (\pi n/L)^2$. The generator, stability guarantee, training protocol and readout are unchanged; since DCT coefficients are real, the whole model is real-valued. Data: $1200$ trajectories per setting, $64 \times 64$, $T = 30$, generated by a cosine pseudo-spectral RK4 solver with the parameters of Appendix~\ref{app:fhn}, at $D_u = 0.01$ and $0.001$ with $D_v = 0.005$ fixed.

\begin{table}[!h]
\caption{FitzHugh--Nagumo with no-flux boundaries.}
\label{tab:noflux}
\centering
\begin{tabular}{lcc}
\toprule
 & $D_u = 0.01$ & $D_u = 0.001$ \\
\midrule
Test relative $L_2$ (3 seeds) & 0.0017 / 0.0017 / 0.0017 & 0.0033 / 0.0034 / 0.0035 \\
max $\mathrm{Re}(\lambda)$ & $-0.084$ & $-0.082$ \\
\bottomrule
\end{tabular}
\end{table}

Accuracy and stability carry over without modification: test error is $0.0017$ and $0.0034$ with every eigenvalue stable, and the learned dissipation slope is larger at the larger diffusivity. The same construction applies to any basis that diagonalizes the dissipative operator under the domain's boundary conditions (cosine for no-flux, sine for absorbing). Chebyshev bases for general Dirichlet conditions and three-dimensional transforms are not tested here.

\section{Optical dispersion: data, readout, and limits}\label{app:nlse}

\paragraph{Data.}
Fibers with $\beta_2 = -5$ to $-30$~ps$^2$/km ($\gamma = 1.3$~W$^{-1}$km$^{-1}$, $\alpha = 0.2$~dB/km, $\beta_3 = 0.1$~ps$^3$/km), integrated by split-step Fourier over 5~km. $1200$ pulses per fiber ($1000/200$), same pulses at every $\beta_2$. Peak power $0.05$--$1$~W, pulse width $T_0 = 2$--$8$~ps, with added random chirp, offset and shape (sech, Gaussian, dual). Fields on 256 points over 100~ps, stored every $\Delta z = 0.166$~km; the model uses the first 20 snapshots (10 in, 10 out), $M = 16$, decoder outputs $[\mathrm{Re}\,E, \mathrm{Im}\,E]$, batch 20; otherwise as Appendix~\ref{app:training}. Test relative $L_2$ is $0.063$--$0.077$ over the 15 models.

\begin{table}[!h]
\caption{$|\beta_2|$ read from the top rotation rate with the derived constant $13.56$, per seed; mean error and standard deviation over three seeds.}
\label{tab:nlse}
\centering
\begin{tabular}{cccccc}
\toprule
$|\beta_2|$ & seed 0 & seed 1 & seed 2 & mean (error) & SD \\
\midrule
10 & 9.67 & 10.24 & 9.20 & 9.70 ($-3.0\%$) & 0.53 \\
15 & 14.67 & 14.20 & 15.18 & 14.68 ($-2.1\%$) & 0.49 \\
20 & 19.95 & 18.76 & 19.42 & 19.38 ($-3.1\%$) & 0.60 \\
25 & 25.41 & 25.05 & 24.39 & 24.95 ($-0.2\%$) & 0.52 \\
30 & 29.36 & 29.97 & 28.86 & 29.40 ($-2.0\%$) & 0.56 \\
\bottomrule
\end{tabular}
\end{table}

\paragraph{Readout.}
Because $S$ is real and skew-symmetric, it has 16 pairs of purely imaginary eigenvalues and their magnitudes are the rotation rates, in radians per snapshot. Projecting each test pulse's latent Fourier coefficient at $\Omega_k$ onto the eigenvectors of $S$ gives the share of mode $k$'s energy on each rotation rate (Fig.~\ref{fig:nlse}a--c, seed 1). With $\Omega_{15} = 2\pi \cdot 15/100~\mathrm{ps} = 0.9425$~rad/ps and $\Delta z = 0.166$~km, $2/(\Omega_{15}^2 \Delta z) = 13.56$.

\paragraph{Limits.}
The band-edge phase advance $|\beta_2|\Omega_{15}^2 \Delta z/2$ must lie between the ${\approx}0.5$~rad occupied by the non-dispersive rotation rates and $\pi$. At $\beta_2 = -5$ it is $0.37$~rad and the readout is $31\%$ high (this bound was set after seeing that point). A 10~km fiber with $\Delta z = 0.332$~km sits at $0.94\pi$ and reads $0.89$ of the true value. Pulse ensembles that do not reach the band edge shift the estimate; $T_0 = 5$--$10$~ps gives $-22\%$, $T_0 = 2$--$4$~ps $+18\%$, peak power $0.5$--$1$~W $-8\%$, a fresh pulse draw $-3.5\%$.

\section{Long-horizon accuracy and unseen time resolution}\label{app:longhorizon}

\paragraph{Long-horizon accuracy.}
We used 200 trajectories at $\nu = 10^{-5}$ with the benchmark's forcing and initial-condition statistics (initial enstrophy $0.95$ of the benchmark test split), integrated to $t = 200$ with an integrating-factor RK4 scheme, since the reference scheme was only run to $t \le 20$ at this viscosity. The baselines use benchmark data for training and are tested on this long-horizon data. DMD: exact DMD on pooled training snapshot pairs, rank 32--256; Table~\ref{tab:baselines} reports the best rank (32 at $\nu = 10^{-3}$, 256 at $\nu = 10^{-5}$). LRAN: the encoder, decoder and training loop of Appendix~\ref{app:training} with the generator replaced by an unconstrained flow map $z_{t+1} = A z_t$ with spectral radius $1.05$ ($\nu = 10^{-3}$) and $1.33$--$1.39$ ($\nu = 10^{-5}$). KNO: the authors' implementation (koopmanlab; operator size 32, 16 modes, 500 epochs, three seeds), error computed with the paper metric. All models receive the same 10-frame input.

\begin{table}[!h]
\caption{Per-step relative $L_2$ (200 trajectories; $t = 15, 20$ are inside the training horizon).}
\label{tab:longhorizon}
\centering
\begin{tabular}{lccccccc}
\toprule
$t$ & 15 & 20 & 30 & 40 & 50 & 100 & 200 \\
\midrule
LGN-KM & 0.16 & 0.36 & 1.06 & 1.11 & 1.04 & 1.01 & 1.00 \\
KNO    & 0.12 & 0.24 & 0.84 & 1.19 & 1.20 & 1.11 & 1.03 \\
FNO    & 0.11 & 0.23 & 0.86 & 1.51 & 1.59 & 72.7 & $1.6 \times 10^{7}$ \\
LRAN   & 0.17 & 0.37 & 14.0 & 249 & $7.6 \times 10^{3}$ & $6.4 \times 10^{10}$ & $\infty$ \\
\bottomrule
\end{tabular}
\end{table}

\paragraph{Unseen time resolution.}
We used 200 fresh trajectories at $\Delta t = 0.25$ step (initial enstrophy $0.96$ of the benchmark), same forcing and viscosity. The model trained on the $\Delta t = 1$ grid is queried at the times between frames 11 and 20 with one matrix exponential each, against linear interpolation of its own integer-time predictions and of the true frames which bounds what interpolation can recover.

\begin{table}[!h]
\caption{Mean relative $L_2$ at unseen times.}
\label{tab:finedt}
\centering
\begin{tabular}{lccc}
\toprule
 & $\exp(L_k t)$ & interpolate own frames & interpolate true frames \\
\midrule
all 27 times & 0.190 & 0.200 & 0.135 \\
half-integer times & 0.190 & 0.205 & 0.159 \\
\bottomrule
\end{tabular}
\end{table}

The direct query beats interpolating the model's own frames at every query time, by the largest margin at half-integer offsets.

\section{Sensitivity of the spectral readout to \texorpdfstring{$r$, $M$}{r, M} and input normalization}\label{app:sensitivity}

For each trained generator at $\nu = 10^{-5}$ we fit $\mathrm{Re}(\lambda)$ against $|k|^2$ on every eigenvalue branch and report the median $R^2$ over branches, the $R^2$ of the dominant (least-damped) branch, and the $R^2$ between the normalized singular-value profile of $S$ and that of the paper model ($r = 32$, $M = 12$, raw input).

\begin{table}[!h]
\caption{Spectral readout vs latent dimension $r$, Fourier truncation $M$ and input normalization (at $\nu = 10^{-5}$).}
\label{tab:sensitivity}
\centering
\small
\begin{tabular}{lcccc}
\toprule
Config & max $\mathrm{Re}(\lambda)$ & median branch $R^2$ & dominant branch $R^2$ & SV-profile $R^2$ vs paper \\
\midrule
$r = 4$  & $-0.005$ & 1.000 & 0.999 & $-1.60$ \\
$r = 8$  & $-0.028$ & 0.993 & 0.996 & 0.45 \\
$r = 16$ & $-0.017$ & 0.997 & 0.890 & 0.958 \\
$r = 32$ (paper) & $-0.021$ & 0.986 & 0.920 & 1 \\
$r = 64$ & $-0.016$ & 0.990 & 0.756 & 0.950 \\
$M = 8$  & $-0.016$ & 0.986 & 0.960 & 0.964 \\
$M = 16$ & $-0.035$ & 0.990 & 0.484 & 0.992 \\
normalized input & $-0.006$ & 0.994 & 0.826 & 0.964 \\
\bottomrule
\end{tabular}
\end{table}

All configurations are stable. The $|k|^2$ dependence holds at median branch $R^2$ $0.986$--$1.000$ across a $16\times$ range of $r$, a $2\times$ range of $M$, and both input conventions. The dominant branch is the minimum-decay envelope: at $M = 16$ and $r = 64$ one slowly decaying channel occupies it at most $k$, so its individual fit weakens while the other branches keep the scaling (at $M = 16$, 29 of 32 branches have $R^2 > 0.9$). The singular-value profile of $S$ is preserved for $r \ge 16$ ($R^2 \approx 0.95$ or above); at $r = 4$ and $8$ the profiles have too few entries to compare. The normalized-input row is a 1000-epoch model trained on z-scored data.

\end{document}